\documentclass[11pt]{article}

\usepackage[final]{acl}

\usepackage{times}
\usepackage{latexsym}

\usepackage[T1]{fontenc}

\usepackage[utf8]{inputenc}

\usepackage{microtype}

\usepackage{inconsolata}

\usepackage{graphicx}
\usepackage{multirow}
\usepackage{algorithm}
\usepackage{amsmath}
\usepackage{enumitem}
\usepackage{subfig}
\usepackage{float}
\usepackage{caption}
\usepackage[capitalize]{cleveref}
\usepackage{siunitx}
\usepackage{algpseudocode}
\usepackage{booktabs}
\usepackage{amssymb}
\usepackage{pdfpages}
\usepackage{longtable}
\usepackage{afterpage}
\usepackage[table]{xcolor}
\usepackage{tabularx}

\title{PyraMathBench: Evaluating and Improving \\Mathematical Capability in Large Language Models}

\author{
  \textbf{Zetian Ouyang\textsuperscript{1}},
  \textbf{Linlin Wang\textsuperscript{1}}\thanks{Corresponding author.},
  \textbf{Gerard de Melo\textsuperscript{2}},
  \textbf{Liang He\textsuperscript{1}}
\\
  \textsuperscript{1}East China Normal University \quad
  \textsuperscript{2}Hasso Plattner Institute, University of Potsdam \\
  \small{
    \texttt{\{51265901102\}@stu.ecnu.edu.cn},
    \texttt{\{llwang,lhe\}@cs.ecnu.edu.cn},
    \texttt{gdm@demelo.org}
  }
}

\begin{document}
\maketitle
\begin{abstract}
Despite the pivotal role of numerical reasoning as the cornerstone of mathematical capabilities in large language models (LLMs) across applications, few benchmarks evaluate LLMs by integrating numerical processing and mathematical reasoning, hindering the interpretability of failures in math tasks. We introduce PyraMathBench\footnote{\url{https://github.com/optifine233-ship-it/PyraMathBench}}, a comprehensive hierarchical benchmark with 32,505 questions derived from 7,404 math word problems, spanning 4 key cognitive aspects, 14 subcategories, and 2 modalities. Experiments reveal that LLMs' performance is severely compromised by inadequate numerical computation and weak handling of abstract numerical questions. To address this, we propose the Smart Optimization \& Learning-based VErsatile module (SOLVE) and Interactive Relative Policy Optimization (IRPO), which enhance LLMs' numerical-mathematical synergy via efficient tool calls (fuzzy matching and low-quality call rejection). Comparative experiments show Qwen-2.5 achieves a 5.0 score improvement with SOLVE and IRPO training.
\end{abstract}

\section{Introduction}
\label{sec: Intro}
Numerical reasoning is ubiquitous in scientific research~\cite{spithourakis-riedel-2018-numeracy}, financial analysis~\cite{chen2019numeracy, jiang-etal-2020-learning}, and integral to text understanding~\cite{yuan2023well, sundararaman2020methods}. Despite advances in data and compute, large language models (LLMs) like GPT-4 and Llama still struggle with math tasks~\cite{patel-etal-2021-nlp, zhao2023survey}, largely due to inadequate numerical processing (e.g., flawed number tokenization~\cite{liu2023goat, yuan2023well}) and numerical hallucination~\cite{ji2023survey, chen2023purr}. 

These issues highlight the pressing need to continually enhance the numerical processing abilities of these models, as these capabilities essentially reflect the model's proficiency in integrating numerical and mathematical reasoning, directly impacting its ability to solve real-world problems and engage in abstract thinking~\cite{wei2022chain}. However, current answer-centric benchmarks fail to disentangle numerical processing flaws from mathematical reasoning errors, lacking mechanisms to diagnose failure modes related to numerical competence. While external tools show promise for boosting capabilities ~\cite{schick2023toolformer, shen2023hugginggpt, qin2023toolllm}, this gap also hinders performance~\cite{yuan2024easytool} due to little interpretability. This creates an urgent need for a high-quality benchmark that enables fine-grained assessment of numerical-mathematical synergy.

Current benchmarks predominantly assess the mathematical reasoning abilities of language models through math word problems (MWPs). Datasets like GSM8K~\cite{cobbe2021training} and APE210K~\cite{zhao2020ape210k}, based on elementary-level problems, and benchmarks such as MATH~\cite{hendrycks2021measuring}, ARB~\cite{sawada2023arb}, and FrontierMath~\cite{glazer2024frontiermath}, which involve competition-level problems like the IMO and AMC, are widely used. However, these benchmarks do not fully capture the limitations of LLMs' capabilities. For example, when models provide incorrect answers, it remains unclear whether the failure stems from computational errors or misinterpretation of the question. 

Some efforts, such as LILA~\cite{mishra-etal-2022-lila}, attempt to address this by breaking down tasks into subtasks. Akhtar et al.~\cite{akhtar2023exploring} introduced a framework to probe LLMs' numerical reasoning at various levels. But these frameworks lack cross-task correlations, testing LLMs’ abilities in isolation without exploring how the models’ abilities to solve simpler tasks may influence their performance on more complex tasks. In fact, mathematical ability can be thought of as a hierarchy, akin to a pyramid structure, where complex tasks are broken down into simpler foundational parts. By isolating and evaluating these core tasks, we can better understand the interplay between foundational skills and higher-level reasoning. This hierarchical approach not only assesses complex tasks but also identifies how weaknesses in basic skills can affect the overall performance. 

Based on these, we propose the PyraMathBench (PMB), a comprehensive hierarchical benchmark providing 32,505 questions derived from 7,404 math word problems, covering 4 key cognitive aspects, 14 subcategories, and 2 modalities. Additionally, the subtasks are decomposed from real math word problems rather than generated, enhancing their relevance for real-world applications. PMB also incorporates the compositional relationships between tasks, decomposing 
math problems into smaller, modular subtasks that can be assessed individually and in combination. Using PMB, we evaluate a variety of state-of-the-art (SOTA) LLMs, identifying areas for improvement and offering valuable insights into the factors that influence performance.

In our analysis, a key weakness observed across the LLMs is their limited capacity for abstraction, equation solving, and factual retrieval. To address this issue, we introduce the Smart Optimization \& Learning-based VErsatile module (SOLVE) to enhance the flexibility and adaptability of LLMs to tool calling, as well as a new RL framework Interactive Relative Policy Optimization (IRPO). Our method yields a 5.0\% improvement against the vanilla LLM. 

Our contributions can be summarized as:
\begin{itemize}
    \item We propose the PyraMathBench, a comprehensive hierarchical benchmark that includes 32,505 questions derived from 7,404 math word problems, covering 4 key cognitive aspects, 14 subcategories, and 2 modalities, ensuring a comprehensive evaluation.
    \item We evaluate a variety of SOTA LLMs with this benchmark, identifying areas for improvement and offering valuable insights into the factors that influence performance.
    \item To address the challenges of mathematical reasoning, we introduce SOLVE module and IRPO algorithm, designed to mitigate tool call parsing failures and improve the efficiency of tool calling.
\end{itemize}

\begin{figure*}
  \centering
  \includegraphics[width=\linewidth]{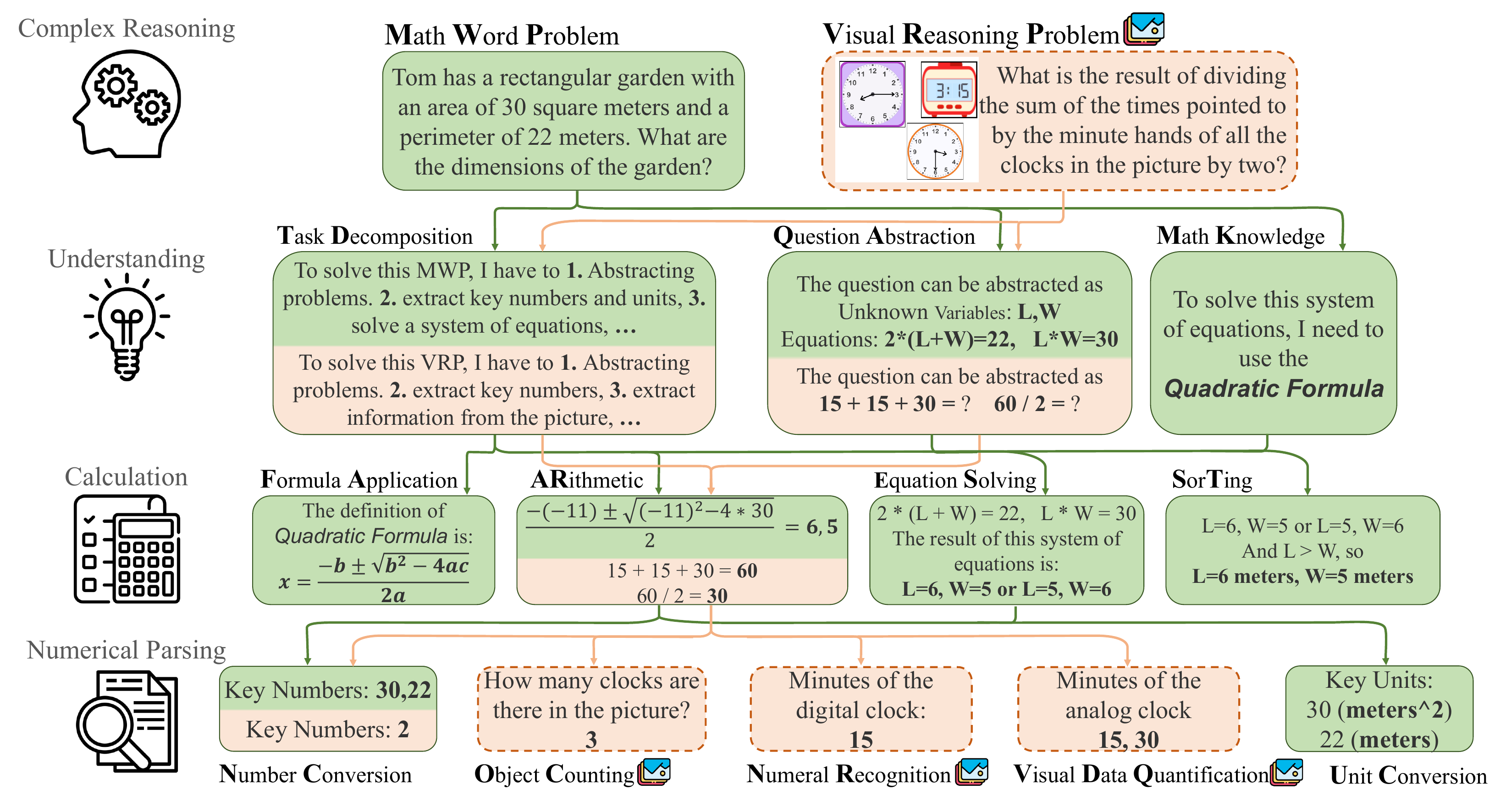}
  \caption{Taxonomy of PyraMathBench and two examples of decomposing complex reasoning problems (green and pink) into subtasks, dashed lines representing multimodal tasks.} 
  \label{fig: framework}
\end{figure*}

\section{The Taxonomy of PyraMathBench}
\label{taxonomy}
The core motivation behind PMB’s taxonomy is the recognition that mathematical tasks often require multiple layers of cognitive aspects and computational skills, ranging from simple numerical parsing to intricate logical reasoning. An LLM's ability to solve a high-level math word problem is contingent upon its proficiency in handling lower-level subcomponents. Previous benchmarks lack an explicit framework for isolating different cognitive aspects, making it difficult to diagnose specific failure points. By decomposing complex mathematical tasks into distinct hierarchical aspects, PMB provides a systematic method to evaluate capability at each stage of mathematical cognition, allowing for a more interpretable assessment of LLM performance.

\begin{figure}
    \centering
    \includegraphics[width=\linewidth]{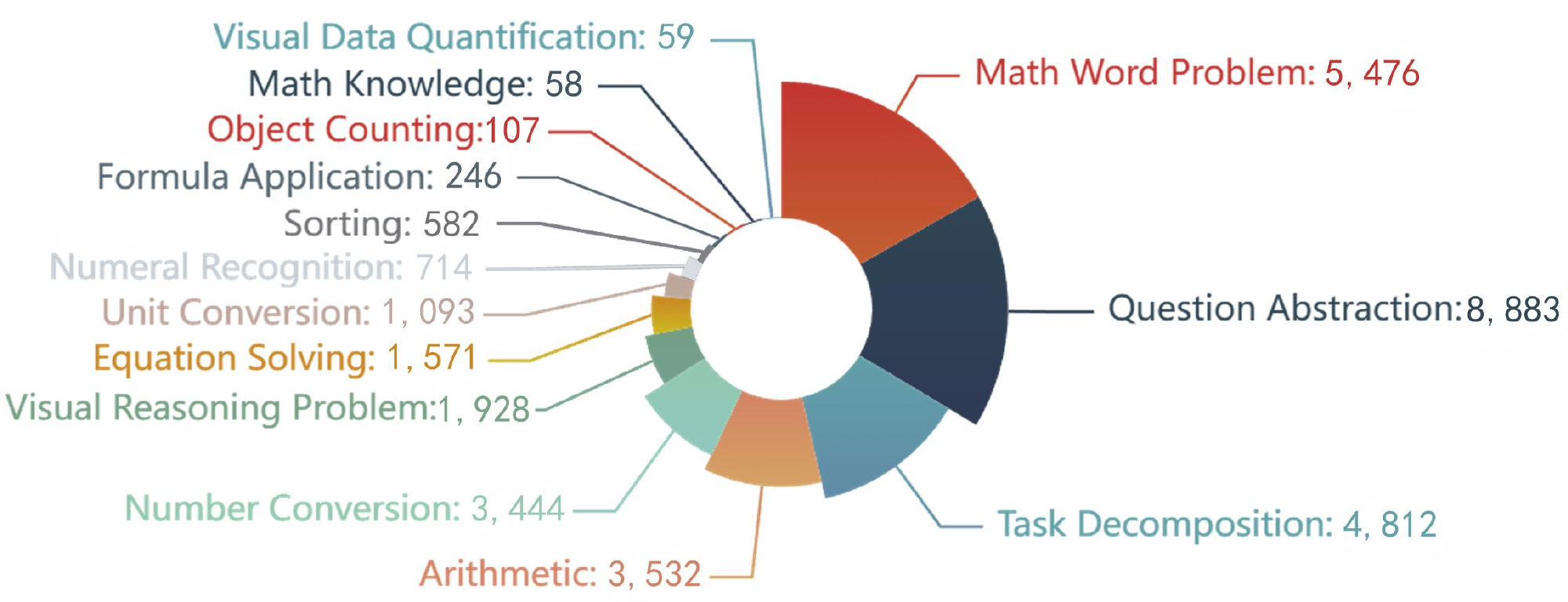}
    \caption{Data distribution of 14 subtasks.}
    \label{fig: subtask distribution}
\end{figure}

Inspired by previous research~\cite{xu2022towards, akhtar2023exploring} and Piaget's cognitive theory, taking into account both the reasoning paradigm and the feasibility of annotation, our benchmark taxonomizes tasks into four hierarchical aspects (A1–A4), encompassing 14 distinct tasks. Figure~\ref{fig: subtask distribution} shows the composition of subtasks at each aspect and examples of subtask annotation. \textbf{Complex Reasoning} evaluates the model’s ability to integrate multiple cognitive processes and mathematical principles. It requires sophisticated logical deductions, image interpretation, and multistep problem-solving. \textbf{Understanding} focuses on the model’s ability to comprehend and interpret mathematical content, transforming unstructured text into mathematical representations.  \textbf{Calculation} primarily tests computational efficiency and correctness in solving mathematical problems. Finally, \textbf{Numerical Parsing} assesses the model’s ability to parse and process numerical data in various formats, thereby evaluating its capacity to recognize, interpret, and extract numerical information for further computation. The specific descriptions, prompts, and examples for 14 subtasks are provided in Appendix~\ref{sec: description of subtasks}.

\section{Construction and Statistics}
\textbf{Data Sources.} The PMB dataset incorporates six existing evaluation datasets and practice question collections. The data collection adheres to the following guidelines: 1) It includes common mathematical problems and visual reasoning tasks to represent the typical problem distribution. 2) Each problem is structured to allow clear decomposition into subtasks, facilitating unambiguous labeling. 3) The dataset is varied in difficulty, ensuring the inclusion of challenging tasks to effectively evaluate the performance of LLMs. We excluded non-mathematical content from the data. Based on this, we considered 6 datasets as data sources: ASDiv~\cite{miao2020diverse}, alg514~\cite{kushman-etal-2014-learning}, Dolphin 18K~\cite{shi-etal-2015-automatically}, SVAMP~\cite{patel-etal-2021-nlp}, TAT-QA~\cite{zhu-etal-2021-tat}, and MathVista~\cite{lu2023mathvista}. Another feature of these datasets is that they provide well-structured answer inference processes or automated question generation tools, facilitating the extraction of subtask questions.

\noindent \textbf{Subtasks Annotation.} The dataset annotation is conducted by six experts proficient in high school-level mathematics. The subtask questions are evenly distributed among the six experts for annotation. For validation the answers annotated by different experts are evaluated using the metrics outlined in Section~\ref{sec: Models and Evaluation Metrics}. If the score falls below 90, the question is deemed ambiguous and subsequently discarded. We also utilized the tabular data from TAT-QA to create images to expand the variety of multimodal tasks.

Additionally, we standardize the mathematical representations across different datasets, ensuring compatibility with both Python interpreters and LaTeX (the latter being used for more complex expressions). For floating-point answers, numerical values are rounded to six decimal places. The detailed data construction process can be found in Appendix~\ref{sec: data construction appendix}.

\noindent \textbf{Statistics.} Figure~\ref{fig: framework} presents the distribution of subtasks. PyraMathBench offers several advantages over existing evaluation methods: (1) \textbf{Comprehensive Coverage} -- PMB includes a diverse array of tasks, spanning four primary areas of mathematical reasoning and 14 subcategories, derived from 32,505 questions across 7,404 Math Word Problems. This extensive dataset facilitates a thorough assessment of models across a wide range of topics and difficulty levels, ensuring broad coverage of mathematical challenges. (2) \textbf{Compositionality of Subtasks} -- PMB structures subtasks derived from the same Math Word Problem, allowing for detailed performance analysis. This compositional approach enables the isolation and evaluation of a model's ability to break down complex problems into simpler components, providing insights into foundational skill deficiencies and their impact on overall performance. (3) \textbf{Multimodal Tasks} -- By incorporating both unimodal and multimodal tasks, PMB enables a more comprehensive evaluation of LLMs. This allows assessing models' ability to process different input types and engage in complex forms of reasoning.

\section{Evaluation}
\label{sec: Models and Evaluation Metrics}
We conduct an evaluation of 11 representative LLMs using the PMB dataset, including GPT-4o, GPT-4o-mini\footnote{https://platform.openai.com/docs/models}, Claude-opus-4.6\footnote{https://www.anthropic.com/api}, LLaVA 13B~\cite{liu2023llava}, DeepSeek-R1~\cite{deepseekai2025deepseekr1incentivizingreasoningcapability}, DeepSeek-Math 7B~\cite{shao2024deepseekmath}, Qwen-2.5 14B~\cite{qwen2.5}, Qwen-2.5-Math 7B~\cite{yang2024qwen2}, Llama3.1 8B~\cite{grattafiori2024llama3herdmodels}, Gemma2 9B~\cite{team2024gemma}, and Mistral 7B~\cite{jiang2023mistral}. To enhance the robustness and reliability of the results with sampling-based decoding, we sample three outputs for each question and calculate the median and deviation of the scores. To simulate real-world mathematical question-answering scenarios, we employed zero-shot settings with Chain of Thought (CoT) prompting~\cite{wei2022chain}. The detailed experiment settings are provided in Appendix~\ref{sec: experiment settings}.

\vspace{-10px}
\begin{equation}
\label{eq: number compare}
\resizebox{\columnwidth}{!}{$\displaystyle
\text{Score}(y, \hat{y}) = \begin{cases}
100 & \text{if\ }\left | \hat{y} - y  \right | < 10^{-4} \\
  0 & \text{if\ }\hat{y}\ \text{is}\ \textsc{Undefined} \\
  \max\!\left(0,\, 100 - \dfrac{\left | \hat{y} - y \right | \cdot 10^{4}}{\max(1, y, \hat{y})}\right) & \text{otherwise}
\end{cases}$}
\end{equation}

The answer types include four formats: 1) a number or list of numbers, 2) expressions, 3) brief text, and 4) multiple-choice options. Numerical answers are evaluated using Equation~\ref{eq: number compare}, where \(y\) represents the reference answer and \(\hat{y}\) represents the model response. The equation is adapted from the relative error definition of~\citet{yuan2023well} and transforms the relative error into a 0--100 score such that a relative error of 1\% yields a score of 0, enabling us to distinguish answer deviations stemming from numerical precision from completely erroneous reasoning. We employ Math-Verify\footnote{https://github.com/huggingface/Math-Verify} to handle diverse mathematical expressions, and normalize structurally different but mathematically equivalent forms (e.g., factored vs.\ expanded, fractional vs.\ decimal) prior to embedding comparison. For text answers, we embed the response via MathBERT~\cite{shen2021mathbert} and calculate the cosine similarity with the answer; the 0.9 matching threshold was calibrated on a held-out 30\% subset of open-ended text questions and achieves near-optimal F1 against expert annotations, with robustness verified against human judgment (98.5\% accuracy, Pearson $r=0.94$; see Appendix~\ref{sec: metric validation}). For multiple-choice questions, we calculate the perfect match rate. Finally, all scores are normalized to a range of 0 to 100. All reported results reflect model versions current as of the dates listed in Appendix~\ref{sec: experiment settings}.

 \begin{table*}[ht] 
 \small
 \centering
 \renewcommand\arraystretch{1}
 \begin{tabular}{l|c|ccccc}
 \hline
 \rowcolor{gray!20}
 \multicolumn{1}{l|}{\textbf{\quad Model}} & \multicolumn{1}{c|}{ \textbf{Size}} & \multicolumn{1}{c|}{MWP} & \multicolumn{1}{c}{QA} & \multicolumn{1}{c}{TD}& \multicolumn{1}{c|}{MK}& \multicolumn{1}{c}{Arithmetic}\\
 \hline
 \rowcolor{gray!20}
 \multicolumn{2}{c|}{\textbf{Aspect}} & \multicolumn{1}{c|}{Reasoning} & \multicolumn{3}{c|}{Understanding}& \multicolumn{1}{c}{Calculation}\\
 \hline
\verb|GPT-4o| & - & \multicolumn{1}{c|}{92.1\(\pm\)0.9} & 66.1\(\pm\)1.6 & 66.6\(\pm\)1.9 & \multicolumn{1}{c|}{75.6\(\pm\)2.1} & 96.0\(\pm\)0.4\\ 
\verb|GPT-4o mini| & - & \multicolumn{1}{c|}{89.6\(\pm\)2.4} & 64.7\(\pm\)2.2 & 61.2\(\pm\)2.4 & \multicolumn{1}{c|}{80.8\(\pm\)1.3} & 95.7\(\pm\)0.9\\ 
\verb|LLaVA| & 13B & \multicolumn{1}{c|}{34.4\(\pm\)0.8} & 11.4\(\pm\)1.7 & 36.1\(\pm\)2.4 & \multicolumn{1}{c|}{61.0\(\pm\)1.8} & 60.9\(\pm\)0.7\\ 
\verb|DeepSeek-R1| & 671B & \multicolumn{1}{c|}{\textbf{93.1\(\pm\)1.1}
} & \textbf{87.7\(\pm\)2.6} & 75.6\(\pm\)1.6 & \multicolumn{1}{c|}{\textbf{100\(\pm\)0}} & \textbf{96.2\(\pm\)1.1}\\ 
\verb|DeepSeek-Math| & 7B & \multicolumn{1}{c|}{59.3\(\pm\)0.7} & 34.7\(\pm\)0.6 & 54.7\(\pm\)0.7 & \multicolumn{1}{c|}{40.1\(\pm\)6.8} & 84.8\(\pm\)0.4\\ 
\verb|Qwen-2.5| & 14B & \multicolumn{1}{c|}{91.1\(\pm\)2.5
} & 64.5\(\pm\)2.7 & 62.7\(\pm\)2.9 & \multicolumn{1}{c|}{83.5\(\pm\)3.0} & 90.4\(\pm\)0.7\\ 
\verb|Qwen-2.5-Math| & 7B & \multicolumn{1}{c|}{82.5\(\pm\)0.6
} & 48.4\(\pm\)2.5 & 45.9\(\pm\)1.3 & \multicolumn{1}{c|}{19.2\(\pm\)1.6} & 95.1\(\pm\)0.3\\ 
\verb|Llama3.1| & 8B & \multicolumn{1}{c|}{72.3\(\pm\)1.1
} & 54.1\(\pm\)2.2 & 76.7\(\pm\)0.1 & \multicolumn{1}{c|}{55.2\(\pm\)7.9} & 89.3\(\pm\)0.4\\
\verb|Gemma2| & 9B & \multicolumn{1}{c|}{87.4\(\pm\)0.8
} & 53.1\(\pm\)1.3 & 77.7\(\pm\)0.6 & \multicolumn{1}{c|}{78.3\(\pm\)2.5} & 81.4\(\pm\)0.9\\
\verb|Mistral| & 7B & \multicolumn{1}{c|}{42.3\(\pm\)1.4
} & 6.6\(\pm\)1.0 & 78.8\(\pm\)0.3 & \multicolumn{1}{c|}{71.0\(\pm\)3.6} & 66.5\(\pm\)0.1\\
\verb|Claude-opus-4.6| & - & \multicolumn{1}{c|}{91.8\(\pm\)4.2} & 87.4\(\pm\)3.4 & \textbf{97.8\(\pm\)0.6} & \multicolumn{1}{c|}{84.6\(\pm\)0} & 95.1\(\pm\)1.2\\
 \hline
 \hline
 \rowcolor{gray!20}
 \multicolumn{1}{l|}{\textbf{\quad Model}} & \multicolumn{1}{c|}{ \textbf{Size}} & \multicolumn{1}{c}{ES} & \multicolumn{1}{c}{Sorting}& \multicolumn{1}{c|}{FA} & \multicolumn{1}{c}{NC} & \multicolumn{1}{c}{UC}\\
 \hline
 \rowcolor{gray!20}
 \multicolumn{2}{c|}{\textbf{Aspect}} & \multicolumn{3}{c|}{Calculation} & \multicolumn{2}{c}{Numerical Parsing}\\
 \hline
\verb|GPT-4o| & - & 97.5\(\pm\)1.1 & 95.8\(\pm\)3.7 & \multicolumn{1}{c|}{76.7\(\pm\)1.3} & \textbf{83.8}\(\pm\)2.7 & 69.7\(\pm\)5.6\\ 
\verb|GPT-4o mini| & - & 98.3\(\pm\)0.5 & \textbf{96.6\(\pm\)3.4} & \multicolumn{1}{c|}{60.5\(\pm\)0.5} & 72.8\(\pm\)2.9 & 73.5\(\pm\)2.4\\ 
\verb|LLaVA| & 13B & 21.6\(\pm\)5.1 & 55.2\(\pm\)1.4 & \multicolumn{1}{c|}{34.6\(\pm\)3.5} & 65.1\(\pm\)0.5 & 29.5\(\pm\)1.9\\ 
\verb|DeepSeek-R1| & 671B & \textbf{99.5\(\pm\)0.5} & 97.4\(\pm\)2.6 & \multicolumn{1}{c|}{\textbf{100\(\pm\)0}} & 81.9\(\pm\)1.1 & 78.6\(\pm\)2.6\\ 
\verb|DeepSeek-Math| & 7B & 90.6\(\pm\)2.9 & 64.6\(\pm\)2.9 & \multicolumn{1}{c|}{21.5\(\pm\)0.6} & 74.9\(\pm\)2.7 & 55.4\(\pm\)2.0\\ 
\verb|Qwen-2.5| & 14B & 98.6\(\pm\)0.5 & 96.0\(\pm\)1.5 & \multicolumn{1}{c|}{79.4\(\pm\)0.7} & 80.9\(\pm\)0.8 & 36.6\(\pm\)1.5\\ 
\verb|Qwen-2.5-Math| & 7B & 97.4\(\pm\)0.7 & 91.5\(\pm\)1.6 & \multicolumn{1}{c|}{77.3\(\pm\)0.9} & 9.7\(\pm\)0.3 & 3.5\(\pm\)0.3\\ 
\verb|Llama3.1| & 8B & 86.8\(\pm\)1.1 & 81.2\(\pm\)2.5 & \multicolumn{1}{c|}{59.5\(\pm\)0.8} & 72.7\(\pm\)0.9 & 24.3\(\pm\)1.7\\
\verb|Gemma2| & 9B & 95.6\(\pm\)0.1 & 93.9\(\pm\)1.4 & \multicolumn{1}{c|}{71.7\(\pm\)1.0} & 69.0\(\pm\)0.6 & 14.9\(\pm\)1.1\\
\verb|Mistral| & 7B & 21.9\(\pm\)1.1 & 90.4\(\pm\)2.0 & \multicolumn{1}{c|}{51.7\(\pm\)0.2} & 64.3\(\pm\)1.4 & 53.9\(\pm\)2.6\\
\verb|Claude-opus-4.6| & - & 99.1\(\pm\)0.9 & 94.7\(\pm\)5.3 & \multicolumn{1}{c|}{86.7\(\pm\)0} & 69.8\(\pm\)4.8 & \textbf{82.7\(\pm\)4.6}\\
 \hline
 \end{tabular}

 \caption{Main results of 11 LLMs on text-only subtasks of PyraMathBench.}
 \label{tab: table1}
\end{table*}

\begin{table}[]
    \small
    \centering
    \setlength{\tabcolsep}{4pt}
    \begin{tabular*}{\columnwidth}{l|c|ccc}
        \hline
        \textbf{Model} & \textbf{VRP} & \textbf{NR} & \textbf{VDQ} & \textbf{OC} \\
        \hline
        \textbf{Aspect} & Reasoning & \multicolumn{3}{c}{Numerical Parsing} \\
        \hline
        GPT-4o          & 73.0\(\pm\)0.7 & 12.3\(\pm\)3.2 & 6.6\(\pm\)2.2 & 1.9\(\pm\)1.8 \\
        GPT-4o mini     & 68.3\(\pm\)1.6 & 14.6\(\pm\)5.6 & 16.7\(\pm\)4.3 & 0\(\pm\)0 \\
        LLaVA           & 25.5\(\pm\)2.8 & 2.8\(\pm\)1.2 & 8.5\(\pm\)2.5 & 7.0\(\pm\)3.1 \\ 
        \hline
    \end{tabular*}
    \caption{Results of 3 MLLMs on multi-modal subtasks of PyraMathBench.}
    \label{tab:placeholder}
\end{table}

\section{Results}
\label{sec: results}
In this section, we analyzed the performance of 11 LLMs on PMB, more analyze through case studies can be found in Appendix~\ref{sec: model analysis}. The results indicate that DeepSeek-R1, Claude-opus-4.6, GPT-4o, and GPT-4o-mini are the top performers, with DeepSeek-R1 exhibiting the highest overall performance. As shown in Table \ref{tab: table1}, DeepSeek-R1 excelled in seven out of eleven text-only tasks, while Claude-opus-4.6 leads in TD and UC. We note that DeepSeek-R1 (671B) is substantially larger than the other open-weight models (7B--14B), so its inclusion is intended as an upper-bound reference for mathematical reasoning rather than a direct head-to-head comparison. Within the comparable closed-source group, Claude-opus-4.6 demonstrated particularly strong understanding and numerical parsing capabilities.

Contrary to expectations, \textbf{math-specific LLMs underperformed on PMB}. Their average score is lower than Llama-3.1, which has a similar number of parameters. This is primarily due to the LLMs' poor instruction-following ability. For example, in the QA and NR subtasks, these models often ignore prompts to extract and abstract from important numerical data, instead directly calculating answers. Additionally, DeepSeek-Math's inconsistent answer formats contributed to its lower performance. These findings suggest that excessive finetuning or reinforcement learning may even hinder an LLM's generalization ability in a specific field.

Furthermore, no MLLM has scored more than 20 on the visual Numerical Parsing subtasks, which is far below their score on the VRP subtask. To assess the statistical reliability of these findings, we conducted bootstrap resampling (1,000 iterations) on GPT-4o across the visual subtasks (NR: 714, VDQ: 59, OC: 107), yielding 95\% confidence intervals that remain well below the VRP performance level, supporting the robustness of the conclusion. This indicates that in fact, \textbf{MLLMs rarely obtain information from images} when solving mathematical image problems and predominantly rely on textual information. Their ability to extract and process mathematical information from images remains fairly underdeveloped, even when the images involved are simple in nature (e.g., counting a certain item in figures or identifying an equation in the diagrams).

In complex reasoning tasks, DeepSeek-R1 achieved the highest score of 93.1 on the MWP task, followed by GPT-4o (92.1) and Claude-opus-4.6 (91.8). Notably, LLaVA shows a rather low performance on the VRP task at 25.5, in contrast to GPT-4o (73.0) and GPT-4o mini (68.3). Overall, \textbf{LLMs exhibited significant weaknesses in the understanding aspect}. Claude-opus-4.6 outperformed the others with an average understanding score of 89.9 (QA: 87.4, TD: 97.8, MK: 84.6), surpassing DeepSeek-R1 (87.8) and far ahead of the third-ranked Qwen-2.5 (70.2). Regarding calculation, the leading LLMs scored around 90--99 on arithmetic and equation-solving tasks. In the Formula Application subtask, DeepSeek-R1 leads with a score of 100, followed by Claude-opus-4.6 (86.7) and Qwen-2.5 (79.4). Notably, Claude-opus-4.6 achieved a near-perfect score of 99.1 on the Equation Solving task and the highest score of 82.7 on the Unit Conversion task, suggesting strong algebraic manipulation and dimensional reasoning capabilities. However, Claude-opus-4.6 underperforms on the Numerical Categorization task (69.8) relative to its overall capability level, falling behind GPT-4o (83.8) and GPT-4o mini (72.8). This contrast indicates that \textbf{Claude-opus-4.6 excels at structured calculation and language understanding, but exhibits relative weakness in categorizing and classifying numerical quantities}, which represents a different capability profile from DeepSeek-R1's more uniformly high performance across tasks. This task requires selecting the correct formula among several variations. The unsatisfactory performance highlights the importance of eliminating hallucinations in mathematical reasoning. The most notable insight regarding Numerical Parsing is the poor performance of MLLMs. However, despite an average score of 9.9 on the Numerical Recognition subtask, the \textbf{MLLMs} do actually recognize a considerable number of digits in the image, but \textbf{fail to determine which digits are useful for solving the problem}, particularly in the presence of redundant data. As a result, MLLMs may also exhibit serious hallucinations in the presence of redundant information in the image. This suggests that previous work~\cite{liu2024chainofspotinteractivereasoningimproves} aimed at improving feature extraction through region-of-interest identification in images may not be sufficiently effective in mathematical contexts.
\begin{figure}[hbtp]
    \centering
    \includegraphics[width=1\linewidth]{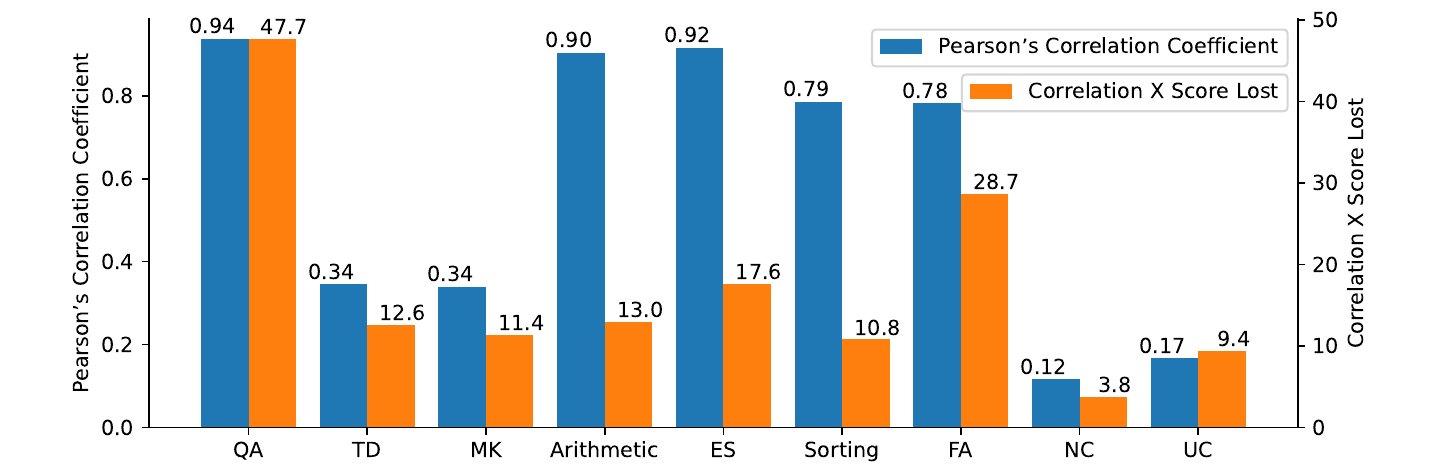}
    \caption{The Pearson’s correlation with MWP and (Correlation X Score Lost) of each subtask.}
    \label{fig: Correlation}
\end{figure}

To quantify the influence of various abilities on performance, we computed the Pearson Correlation Coefficient between MWP scores and each subtask. The results, shown in Figure~\ref{fig: Correlation}, reveal that QA (0.94), ES (0.92), AR (0.90), ST (0.79), and FA (0.78) are strongly correlated with MWP performance. When considering score losses, QA (47.7), ES (17.6), and FA (28.7) emerged as the key weaknesses of LLMs. This analysis suggests that \textbf{LLMs' performance on complex mathematical reasoning tasks is strongly influenced by their ability to handle calculations and abstract questions}. To address these issues, we propose a method in Section~\ref{sec: SOLVE and IRPO} to enhance LLMs' mathematical capabilities through tool calls.

\begin{figure}[hbtp]
    \centering
    \includegraphics[width=\linewidth]{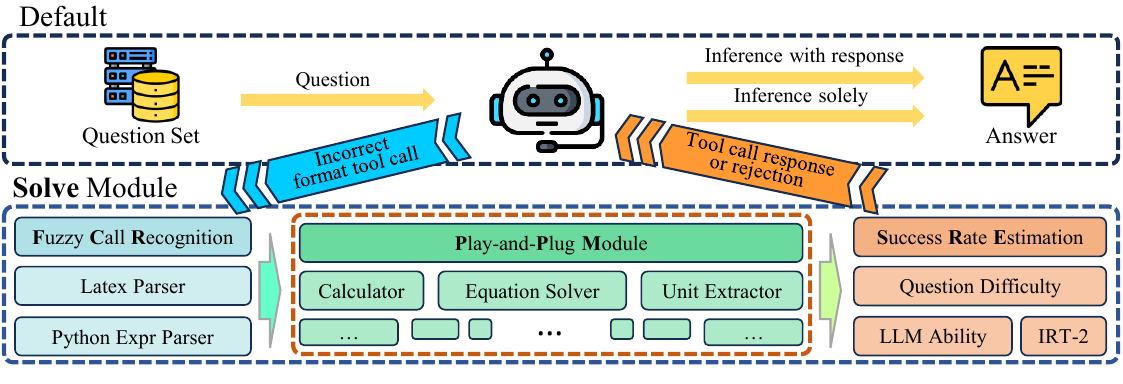}
    \caption{The workflow of SOLVE. The call parsing module is used to extract expressions from tool calls, and the difficulty assess module is used to assess tool call difficulty, when the difficulty is below the threshold, SOLVE will bypass the tool call.}
    \label{fig: FCM workflow}
\end{figure}

{
\begin{figure*}
\small
\begin{equation}
    \begin{aligned}\mathcal{J}_{IRPO}(\theta) & =\mathbb{E}\left[q \sim P(Q),\left\{o_{i}\right\}_{i=1}^{G} \sim \pi_{\theta_{ref}}(O \mid q),\left\{\left\{ o_{i,j} \right\}_{j=1}^{N_i}\right\}_{i=1}^{G}\sim \pi_{\theta_{ref}}^{\tau}(O \mid q)\right] \\& 
    \frac{1}{G} \sum_{i=1}^{G} \left\{\frac{1}{\left|o_{i}\right|} \sum_{t=1}^{\left|o_{i}\right|}\left(\frac{\pi_{\theta}\left(o_{i, t} \mid q, o_{i,<t}\right)}{\pi_{\theta_{ref}}\left(o_{i, t} \mid q, o_{i,<t}\right)} \hat{A}_{i, t}-\beta \mathbb{D}_{K L}\left[\pi_{\theta}| | \pi_{r e f}\right]\right) +\right. \\& \left. \frac{1}{\left|o_{i}\right|}\sum_{t=1}^{\left|o_{i}\right|}\left[\frac{1}{N_{i,t}}\sum_{j=1}^{N_{i,t}}\left(\frac{\pi_{\theta}^{\tau}\left(o_{i, j, t} \mid q, o_{i,j, <t}\right)}{\pi_{ref}^{\tau}\left(o_{i, j, t} \mid q, o_{i,j, <t}\right)} \hat{A}_{i, j, t}-\beta \mathbb{D}_{K L}\left[\pi_{\theta}^{\tau}| | \pi_{ref}^{\tau}\right]\right)\right]\right\}\end{aligned}
    \label{eq: TRPO}
\end{equation}
\end{figure*}
}

\section{The proposed SOLVE module and IRPO}
\label{sec: SOLVE and IRPO}
Enhancing LLMs with external tools has emerged as a promising approach to improve their computational capabilities~\cite{schick2023toolformer, shen2023hugginggpt, qin2023toolllm}. To assess the effectiveness of the tool calling approach in mitigating the issues identified in Section~\ref{sec: results}, we first implemented a simple Play-and-Plug tool calling Module (PPM). Then we compared the accuracy of Qwen-2.5 with and without PPM on the PMB/MWP datasets, as shown in Figure~\ref{fig: FCM workflow}. The results reveal a slight decrease in accuracy when using tools. Upon conducting a case analysis of the model's output, we identified the primary reasons for the increase in error rate as: (1) LLMs failing to generate tool call requests in the instructed format, (2) tool call iterations increasing the context length, posing a challenge to LLM's understanding, and (3) LLMs tending to call tools for simple questions, which were unnecessary for the reasoning process.

To address these issues, we introduce the Smart Optimization \& Learning-based VErsatile module (SOLVE) as well as Interactive Relative Policy Optimization (IRPO), designed to mitigate tool call parsing failures and improve the efficiency of tool calling. The specific implementation can be found in the Appendix~\ref{sec: design of solve module} and Appendix~\ref{sec: implementation of IRPO}, respectively.

\subsection{The Smart Optimization \& Learning-based VErsatile module}
SOLVE enhances the flexibility and adaptability of tool integration to make it superior to simple PPM. Figure~\ref{fig: FCM workflow} illustrates the workflow of SOLVE. While preserving plug-and-play compatibility, SOLVE improves tool call efficiency through fuzzy call recognition, i.e. the LLMsl can call tools in various styles, not limited to a certain format. SOLVE offloads the responsibility of formatting tool call outputs from the LLM, allowing the model to produce unstructured or loosely formatted tool calls. SOLVE then standardizes these outputs expressed in LaTeX, Python syntax, or informal handwritten styles, substantially increasing the success rate of tool invocation. Given that different LLM series' tool call ability may be trained in different formats, this feature becomes particularly important.

Additionally, SOLVE employs a rule-based mechanism to estimate the difficulty and discrimination of tool calls. These parameters are incorporated into a two-parameter logistic Item Response Theory model (IRT-2PL)~\cite{cai2016item} to estimate the probability of the LLM autonomously solving the tool call task correctly, formulated as \(P(X_{ij}=1|\theta_j)=\frac{1}{1+e^{-a_i(\theta-b_i)}}\), where \(a_i\) and \(b_i\) denote the discrimination and difficulty of question $i$, respectively, and \(\theta_j\) represents the model's proficiency, which SOLVE will dynamically update based on recent performance. When \(P(X_{ij}=1|\theta_j)\) exceeds a predefined confidence threshold (0.95 in our experiments), SOLVE bypasses tool calls and delegates the task directly to the LLM, optimizing both reliability and computational efficiency.

\subsection{Interactive Relative Policy Optimization}

\begin{figure}
    \centering
    \includegraphics[width=\linewidth]{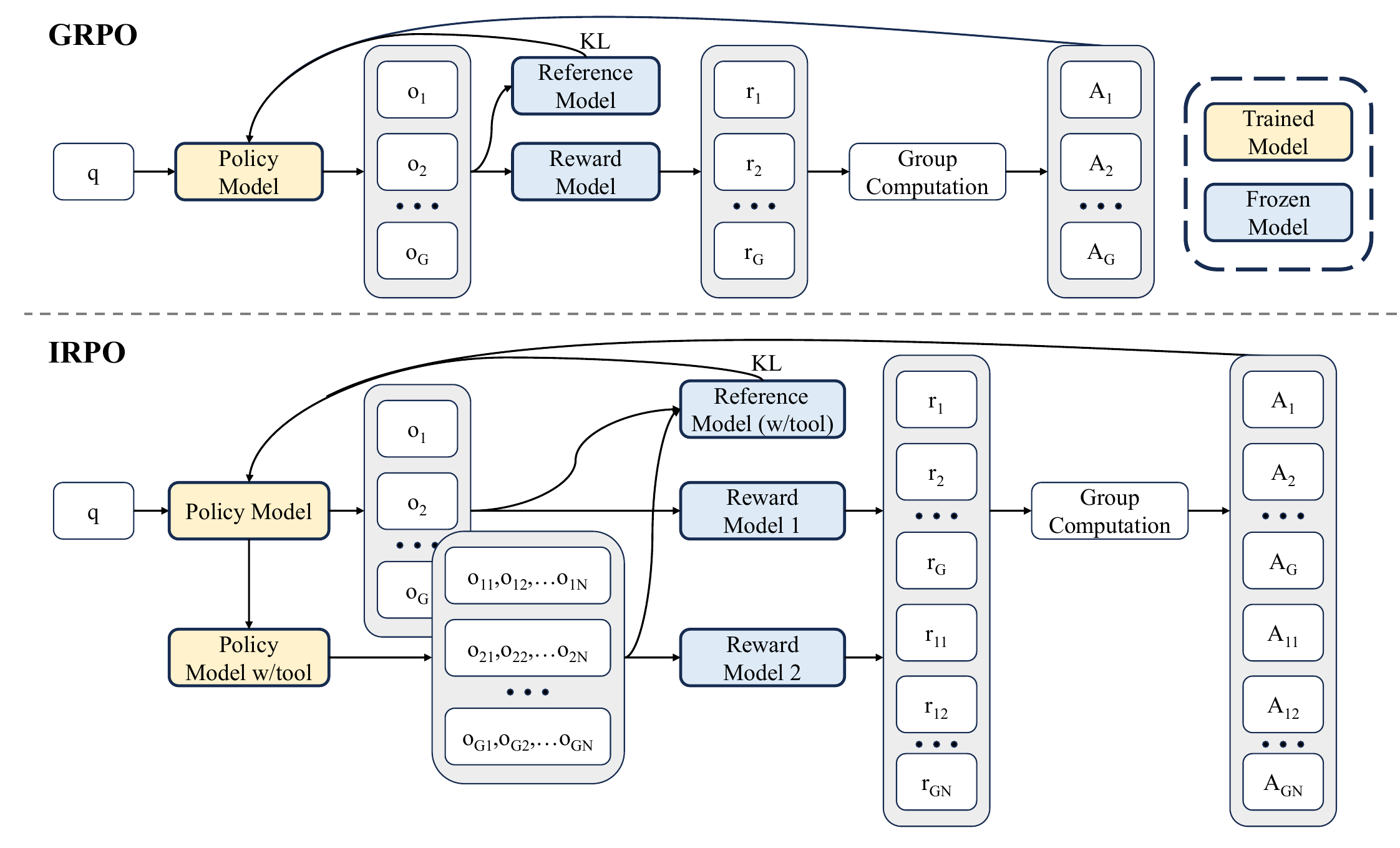}
    \caption{Demonstration of GRPO and our IRPO. IRPO samples two sets of model outputs and calculates the advantages using different reward models/functions.}
    \label{fig: GRPO and TRPO}
\end{figure}

\subsubsection{Optimization Objective}
Reinforcement learning (RL) has demonstrated efficacy in guiding LLMs to align with desired behaviors~\cite{wang2023math, luo2023wizardmath}.  In tool-augmented inference, existing RL approaches like GRPO~\cite{shao2024deepseekmath} optimize the LLM's performance purely from the model's perspective. These methods primarily reward the model based on its answers without considering the involvement of external tools. To address this gap, we propose a new RL framework that extends GRPO by incorporating tool usage, as depicted in Figure~\ref{fig: GRPO and TRPO}. 

For a given question q, the IRPO simultaneously samples two sets of outputs, one from the base policy model \(\pi_{\theta}\), and another from the tool-augmented policy model \(\pi^{\tau}_{\theta}\), which include both intermediate tool-related responses and final LLM responses. The outputs are denoted as \(\left\{o_1, o_2,...,o_G\right\}\) and \(\left\{\left\{o_{11}, o_{12},...,o_{1N_1}\right\},...,\left\{o_{G1}, o_{G2},...,o_{GN_G}\right\}\right\}\), where \(N_i\) is the response round of the i-th response, and \(o_ {ij}, j<N_i \) represents the $j$-th round to the $i$-th response, which includes a tool call request and the result of that request, and \(o_ {ij}, j=N_i \)represents the final response of the LLM when using external tools. The IRPO optimization objective is given by:
where \(\pi_{\theta}\) and \(\pi_{\theta_{ref}}\) represent the current and reference policy models, respectively, and q denote the questions sampled from the reference policy model \(\pi_{\theta_{ref}}\). The KL divergence, \(\mathbb{D}_{K L}\), is estimated using an unbiased estimator~\cite{schulman2020approximating}.

The IRPO objective is tailored for tool-involving scenarios. The advantage of this formulation is that it simultaneously optimizes the model's decision-making ability and the tool invocation strategy. For simplicity, we retain only the KL divergence term to stabilize training and omit the truncation for parameter updates.

\subsubsection{Advantage and Reward Function}
To calculate the advantages for individual output tokens in base and tool-involving scenarios, denoted as \(A_{i,t}\) and \(A_{i,j,t}\), we propose rule-based reward functions to evaluate the model outputs. Specifically, given the model answer \(\hat{y}\) and the reference \(y\), the reward for the final answer is determined by its accuracy, calculated with Equation~\ref{eq: number compare} as \(r_i=\text{Score}(y, \hat{y})\) and \(r_{i,j=N_{i}}=\text{Score}\left(y_{iN_{i}}, \hat{y_i}\right)\). For multi-round tool-call rewards, we use the Equation~\ref{eq: TRPO reward function}:
{
\begin{equation}
    \begin{aligned}
        &r_{i,j<N_{i}} = \frac{1}{3}\left(\mathbb I_{q_{sub}}\left(q_{ij}\right)\times \sqrt{1-\text{Acc}^2\left(q_{ij}\right)}\right. \\
        & + \left.\frac{e^{-\text{Mean}(r_i)}-e^{-1}}{1-e^{-1}}+\frac{1}{1+e^{-a_q(\theta-b_q)}}\right)
    \end{aligned}
    \label{eq: TRPO reward function}
\end{equation}
}
where \(q_{ij}\) represents the function call part of \(o_{ij}\), \(\mathbb I_{q_{sub}}\) is the indicator function of the PMB subquestion set, and \(\text{Acc}(q_{ij})\) is the accuracy of \(q_{ij}\) calculated in the main experiment. The formula has three components:
\begin{itemize}[label=$\bullet$,leftmargin=*]
    \item If the tool call is a PMB subtask, the score decreases with higher accuracy, implying the model can independently solve the problem.
    \item The second term calculates the average score output from the default policy model, where higher scores suggest direct computation without tool calls.
    \item The third part estimates the probability that the LLM with proficiency \(\theta\) generates a correct response. The parameters for this component are stored in the SOLVE module and are dynamically updated across tool invocations. The rewards from this component are dependent on the problem and model proficiency, enabling the SOLVE component to decide when to call a tool based on the model’s capabilities and the problem's complexity.
\end{itemize}

Finally, based on the collected rewards \(\textbf{r}=\left\{r_1, r_2,...,r_G,r_{11}, r_{12},...,r_{GN_G}\right\}\), we normalize the output tokens in each round as the advantages \(\hat{A}_{ij}=\widetilde{r_i}=\frac{r_i-\text{Mean}(\textbf{r})}{\text{Std}(\textbf{r})}\) and \(\hat{A_{i,j,t}}=\widetilde{r_{ij}}=\frac{r_ {ij}-\text{Mean} (\textbf{r})}{\text{Std}(\textbf{r})}\). The policy is then optimized by maximizing the objective function outlined in Equation~\ref{eq: TRPO}. Algorithm~\ref{alg: TRPO algorithm} provides a summary of the IRPO workflow.

\begin{algorithm}[ht]
\small
\caption{IRPO}
\label{alg: TRPO algorithm}
\begin{algorithmic}[1]
\State \textbf{Input:} policy model \(\pi_{\theta_{init}}\); reward functions \(r_{func}\); task prompts \(\mathcal{D}\)
\State \(\pi_{\theta} \leftarrow \pi_{\theta_{init}}\); \(\pi_{\theta_{ref}} \leftarrow \pi_{\theta_{init}}\); \(\pi^{\tau}_{\theta_{ref}} \leftarrow \pi_{\theta_{init}}\)
\For{step = 1,...,M}
    \State Sample a batch \(\mathcal{D_b}\) from \(\mathcal{D}\)
    \State Sample G outputs \(\left\{o_i\right\}^G_{i=1}\sim\pi_{\theta_{ref}}(\cdot|q)\) for each question \(q \in \mathcal{D_b}\)
    \State Sample G outputs \(\left\{o_{i1}, o_{i2},...,o_{iN_{i}}\right\}^G_{i=1}\sim\pi^{\tau}_{\theta_{ref}}(\cdot|q)\) for each question \(q \in \mathcal{D_b}\)
    \State Compute \(\textbf{r}\) for each sample output by \(r_{func}\)
    \State Compute \(\hat{A}_{i,t}\) and \(\hat{A}_{i,j,t}\) for each token
    \For{IRPO iteration = 1,...,\(\mu\)}
        \State Update the policy model \(\pi_{\theta}\)
    \EndFor
\EndFor
\State \textbf{Output:} \(\pi_{\theta}\)
\end{algorithmic}
\end{algorithm}

\begin{table}[ht]
\small
 \centering
 \renewcommand\arraystretch{1}
 \begin{tabular}{llccc}
 \toprule
 \textbf{Model} & \textbf{Training} & No & PPM & SOLVE\\
 \midrule
 \multirow{3}{*}{Qwen-2.5}
 & Default LLM & 91.1 & 82.8 & 95.8\\
 & IRPO & N/A & 90.2 & \textbf{96.1}\\
 & GRPO & 91.3 & 84.3 & 95.5\\
 \midrule
 \multirow{3}{*}{Llama-3.1-8B}
 & Default LLM & 72.3 & 66.5 & 74.8\\
 & IRPO & N/A & 69.0 & \textbf{76.1}\\
 & GRPO & 73.0 & 70.1 & 72.9\\
 \bottomrule
 \end{tabular}
 \caption{Comparison of score on the MWP dataset for Qwen-2.5 and Llama-3.1-8B with different tool call modules and reinforcement learning algorithms.}
 \label{tab: Tool call result}
 \end{table}

\subsection{Experimental Results}
We conduct experiments to evaluate the perfor- mance of Qwen-2.5 and Llama-3.1-8B with different tool call modules and reinforcement learning algorithms. These configurations are assessed on the MWP dataset to gauge their impact on the model's performance in complex reasoning tasks. The experimental results are presented in Table~\ref{tab: Tool call result}.

For both models, the PPM configuration resulted in degraded performance compared to the default LLM baseline (Qwen-2.5: 91.1 $\to$ 82.8; Llama-3.1-8B: 72.3 $\to$ 66.5). Upon examining the output, we identified numerous tool call failures due to issues such as parameter errors and parsing errors, primarily attributed to the challenges of using tool documentation to guide LLMs in tool invocation.

In contrast, the integration of the SOLVE module consistently improved performance across both models (Qwen-2.5: 95.8; Llama-3.1-8B: 74.8). This improvement underscores SOLVE's enhanced flexibility and adaptability through fuzzy call recognition and efficient tool invocation strategies. Further, applying IRPO alongside the SOLVE module led to additional gains, achieving the highest scores for both models (Qwen-2.5: 96.1; Llama-3.1-8B: 76.1). GRPO, however, exhibited inconsistent behavior: while it matched IRPO's trend for Qwen-2.5 (95.5 with SOLVE), it underperformed even the default baseline for Llama-3.1-8B when paired with SOLVE (72.9), falling below the default LLM+SOLVE result (74.8). This suggests that GRPO's inability to model multi-step tool invocation relationships is more pronounced in weaker models, whereas IRPO's explicit multi-round optimization remains effective regardless of model capacity.

\section{Related Work}
The evaluation of LLMs in mathematical reasoning has seen significant advancements through the development of various benchmarks targeting distinct cognitive tasks and problem-solving abilities. MWPs have been a central focus, as they mirror real-world applications of mathematical reasoning and knowledge integration. Datasets like GSM8K~\cite{cobbe2021training}, APE210K~\cite{zhao2020ape210k}, MATH401~\cite{yuan2023well}, and Math23K~\cite{wang2017deep} provide diverse problem sets ranging from elementary to undergraduate levels, assessing foundational to advanced reasoning skills. In pursuit of more rigorous assessments, the Advanced Reasoning Benchmark~\cite{sawada2023arb} sourced from graduate-level exams and professional resources, covering topics from undergraduate to early graduate curricula. OlympiadBench~\cite{he2024olympiadbench}, FrontierMath~\cite{glazer2024frontiermath}, PutnamBench~\cite{tsoukalas2024putnambench}, and OmniMATH~\cite{gao2024omni} focus on olympiad-level mathematics, curating problems from international competitions like IMO and AMC.  However, these benchmarks do not fully capture the limitations of LLMs' capabilities. For example, when models provide incorrect answers, it remains unclear whether the failure stems from computational errors or misinterpretation of the question.  Some efforts, such as LILA~\cite{mishra-etal-2022-lila}, attempt to address this by breaking down tasks into subtasks. Akhtar et al.~\cite{akhtar2023exploring} introduced a framework to probe LLMs' numerical reasoning at various levels. However, these frameworks lack cross-task correlations, testing LLMs’ abilities in isolation without exploring how the models’ abilities to solve simpler tasks may influence their performance on more complex tasks.

The mathematical ability of MLLMs is also a focus in both academia and industry, MathVista~\cite{lu2023mathvista} is a benchmark designed to combine challenges from diverse mathematical and visual tasks and systematically analyze the mathematical reasoning capabilities of SOTA MLLMs in visually complex scenarios. MathVerse~\cite{zhang2024mathverse} meticulously collects 2,612 high-quality, multi-subject math problems with diagrams to assess whether and how much MLLMs can truly understand the visual diagrams for mathematical reasoning. However, these evaluation benchmarks take two modalities as inputs and assess the model's image understanding ability, ignoring the impact of textual information on the model's accuracy, as better answers may be due to sufficient information provided in the text.

In terms of equipping LLMs with tools, two primary approaches have been explored: (1) fine-tuning open-source LLMs~\cite{schick2023toolformer, qin2023toolllm, patil2024gorilla, parisi2022talm, hao2023toolkengpt}, and (2) providing tool documentation and few-shot demonstrations~\cite{shen2023hugginggpt, song2023restgpt, lu2023chameleon, xu2023tool}. However, these methods either introduce additional data collection and model training overhead or are limited by issues such as document inconsistency, redundancy, and incompleteness, which hinder effective tool utilization~\cite{yuan2024easytool}. Hsieh et al.~\cite{hsieh2023tool} proposed a modified tool documentation method to support zero-shot tool calling, which, however, did not resolve the inherent challenges of tool calling. Goldie et al.~\cite{goldie2025synthetic} proposed a synthetic data generation and RL methodology targeting multi-step optimization scenarios, but without aligning each round of tool calling, and not specifically designed for mathematics.

\section{Conclusion}
This paper proposes PyraMathBench, a comprehensive hierarchical benchmark that includes 32,505 questions derived from 7,404 math word problems, covering 4 key cognitive aspects, 14 subcategories, and 2 modalities. Our evaluation of multiple LLMs and MLLMs highlights their limitations in question abstraction, equation solving, and image-based information extraction, which impede accurate inferences on complex mathematical tasks. We also propose the SOLVE module and IRPO algorithm, both designed to mitigate tool call parsing failures and improve the efficiency of tool calling. Experimental results demonstrate that this model significantly improves LLMs' performance in solving math word problems.

\section{Limitations}
This study annotates subtasks by decomposing the MWP and VRQ problems, though it is important to note that there could be multiple versions of decomposition regarding task types and content. While various strategies have been employed to mitigate the impact of this issue during evaluation(e.g., arithmetics and univariate equations share the same evaluation status), it might still influence the results, particularly in the \textbf{Understanding} aspect. Furthermore, our task decomposition method does not independently evaluate the full range of LLM language capabilities except for the 14 subtasks, which means our taxonomy does not include all possible atomic tasks. This is a direction for our future work. Moreover, the current study focuses on English only. Additional research could be conducted on a diverse range of further languages.

While the SOLVE module and IRPO are designed to enhance LLMs' performance on MWP-related tasks, their effectiveness may not be as pronounced in other mathematical domains, such as formula proofs or algebraic calculations, which have a fundamentally different solution path from MWP.

\section*{Acknowledgments}
This work was supported by the Computational Biology Program (Grant No. 25JS2830400 \& 25JS2830402) of Science and Technology Commission of Shanghai Municipality, and Shanghai Municipal Science and Technology Major Project (Grant No. 2025SHZDZX025G06).

\bibliography{custom}

@inproceedings{patel-etal-2021-nlp,
    title = "Are {NLP} Models really able to Solve Simple Math Word Problems?",
    author = "Patel, Arkil  and
      Bhattamishra, Satwik  and
      Goyal, Navin",
    editor = "Toutanova, Kristina  and
      Rumshisky, Anna  and
      Zettlemoyer, Luke  and
      Hakkani-Tur, Dilek  and
      Beltagy, Iz  and
      Bethard, Steven  and
      Cotterell, Ryan  and
      Chakraborty, Tanmoy  and
      Zhou, Yichao",
    booktitle = "Proceedings of the 2021 Conference of the North American Chapter of the Association for Computational Linguistics: Human Language Technologies",
    month = jun,
    year = "2021",
    address = "Online",
    publisher = "Association for Computational Linguistics",
    url = "https://aclanthology.org/2021.naacl-main.168/",
    doi = "10.18653/v1/2021.naacl-main.168",
    pages = "2080--2094"
}

@article{zhao2023survey,
  title={A survey of large language models},
  author={Zhao, Wayne Xin and Zhou, Kun and Li, Junyi and Tang, Tianyi and Wang, Xiaolei and Hou, Yupeng and Min, Yingqian and Zhang, Beichen and Zhang, Junjie and Dong, Zican and others},
  journal={arXiv preprint arXiv:2303.18223},
  year={2023}
}

@article{liu2023goat,
  title={{GOAT}: Fine-tuned {LLaMA} Outperforms {GPT-4} on Arithmetic Tasks},
  author={Liu, Tiedong and Low, Bryan Kian Hsiang},
  journal={arXiv preprint arXiv:2305.14201},
  year={2023}
}

@article{yuan2023well,
  title={How well do large language models perform in arithmetic tasks?},
  author={Yuan, Zheng and Yuan, Hongyi and Tan, Chuanqi and Wang, Wei and Huang, Songfang},
  journal={arXiv preprint arXiv:2304.02015},
  year={2023}
}

@inproceedings{sundararaman2020methods,
  title={Methods for numeracy-preserving word embeddings},
  author={Sundararaman, Dhanasekar and Si, Shijing and Subramanian, Vivek and Wang, Guoyin and Hazarika, Devamanyu and Carin, Lawrence},
  booktitle={Proceedings of the 2020 Conference on Empirical Methods in Natural Language Processing (EMNLP)},
  pages={4742--4753},
  year={2020}
}

@inproceedings{spithourakis-riedel-2018-numeracy,
    title = "Numeracy for Language Models: Evaluating and Improving their Ability to Predict Numbers",
    author = "Spithourakis, Georgios  and
      Riedel, Sebastian",
    editor = "Gurevych, Iryna  and
      Miyao, Yusuke",
    booktitle = "Proceedings of the 56th Annual Meeting of the Association for Computational Linguistics (Volume 1: Long Papers)",
    month = jul,
    year = "2018",
    address = "Melbourne, Australia",
    publisher = "Association for Computational Linguistics",
    url = "https://aclanthology.org/P18-1196/",
    doi = "10.18653/v1/P18-1196",
    pages = "2104--2115"
}

@article{ji2023survey,
  title={Survey of hallucination in natural language generation},
  author={Ji, Ziwei and Lee, Nayeon and Frieske, Rita and Yu, Tiezheng and Su, Dan and Xu, Yan and Ishii, Etsuko and Bang, Ye Jin and Madotto, Andrea and Fung, Pascale},
  journal={ACM Computing Surveys},
  volume={55},
  number={12},
  pages={1--38},
  year={2023},
  publisher={ACM New York, NY}
}

@article{chen2023purr,
  title={Purr: Efficiently editing language model hallucinations by denoising language model corruptions},
  author={Chen, Anthony and Pasupat, Panupong and Singh, Sameer and Lee, Hongrae and Guu, Kelvin},
  journal={arXiv preprint arXiv:2305.14908},
  year={2023}
}

@article{wei2022chain,
  title={Chain-of-thought prompting elicits reasoning in large language models},
  author={Wei, Jason and Wang, Xuezhi and Schuurmans, Dale and Bosma, Maarten and Xia, Fei and Chi, Ed and Le, Quoc V and Zhou, Denny and others},
  journal={Advances in neural information processing systems},
  volume={35},
  pages={24824--24837},
  year={2022}
}

@inproceedings{chen2019numeracy,
  title={Numeracy-600K: Learning numeracy for detecting exaggerated information in market comments},
  author={Chen, Chung-Chi and Huang, Hen-Hsen and Takamura, Hiroya and Chen, Hsin-Hsi},
  booktitle={Proceedings of the 57th Annual Meeting of the Association for Computational Linguistics},
  pages={6307--6313},
  year={2019}
}

@inproceedings{jiang-etal-2020-learning,
    title = "Learning Numeral Embedding",
    author = "Jiang, Chengyue  and
      Nian, Zhonglin  and
      Guo, Kaihao  and
      Chu, Shanbo  and
      Zhao, Yinggong  and
      Shen, Libin  and
      Tu, Kewei",
    editor = "Cohn, Trevor  and
      He, Yulan  and
      Liu, Yang",
    booktitle = "Findings of the Association for Computational Linguistics: EMNLP 2020",
    month = nov,
    year = "2020",
    address = "Online",
    publisher = "Association for Computational Linguistics",
    url = "https://aclanthology.org/2020.findings-emnlp.235/",
    doi = "10.18653/v1/2020.findings-emnlp.235",
    pages = "2586--2599"
}

@inproceedings{mishra-etal-2022-lila,
    title = "{LILA}: A Unified Benchmark for Mathematical Reasoning",
    author = "Mishra, Swaroop  and
      Finlayson, Matthew  and
      Lu, Pan  and
      Tang, Leonard  and
      Welleck, Sean  and
      Baral, Chitta  and
      Rajpurohit, Tanmay  and
      Tafjord, Oyvind  and
      Sabharwal, Ashish  and
      Clark, Peter  and
      Kalyan, Ashwin",
    editor = "Goldberg, Yoav  and
      Kozareva, Zornitsa  and
      Zhang, Yue",
    booktitle = "Proceedings of the 2022 Conference on Empirical Methods in Natural Language Processing",
    month = dec,
    year = "2022",
    address = "Abu Dhabi, United Arab Emirates",
    publisher = "Association for Computational Linguistics",
    url = "https://aclanthology.org/2022.emnlp-main.392/",
    doi = "10.18653/v1/2022.emnlp-main.392",
    pages = "5807--5832"
}

@article{cobbe2021training,
  title={Training verifiers to solve math word problems},
  author={Cobbe, Karl and Kosaraju, Vineet and Bavarian, Mohammad and Chen, Mark and Jun, Heewoo and Kaiser, Lukasz and Plappert, Matthias and Tworek, Jerry and Hilton, Jacob and Nakano, Reiichiro and others},
  journal={arXiv preprint arXiv:2110.14168},
  year={2021}
}

@article{sawada2023arb,
  title={Arb: Advanced reasoning benchmark for large language models},
  author={Sawada, Tomohiro and Paleka, Daniel and Havrilla, Alexander and Tadepalli, Pranav and Vidas, Paula and Kranias, Alexander and Nay, John J and Gupta, Kshitij and Komatsuzaki, Aran},
  journal={arXiv preprint arXiv:2307.13692},
  year={2023}
}

@article{glazer2024frontiermath,
  title={Frontiermath: A benchmark for evaluating advanced mathematical reasoning in ai},
  author={Glazer, Elliot and Erdil, Ege and Besiroglu, Tamay and Chicharro, Diego and Chen, Evan and Gunning, Alex and Olsson, Caroline Falkman and Denain, Jean-Stanislas and Ho, Anson and Santos, Emily de Oliveira and others},
  journal={arXiv preprint arXiv:2411.04872},
  year={2024}
}

@article{zhao2020ape210k,
  title={Ape210k: A large-scale and template-rich dataset of math word problems},
  author={Zhao, Wei and Shang, Mingyue and Liu, Yang and Wang, Liang and Liu, Jingming},
  journal={arXiv preprint arXiv:2009.11506},
  year={2020}
}

@inproceedings{hendrycks2021measuring,
  title={Measuring Mathematical Problem Solving With the {MATH} Dataset},
  author={Hendrycks, Dan and Burns, Collin and Kadavath, Saurav and Arora, Akul and Basart, Steven and Tang, Eric and Song, Dawn and Steinhardt, Jacob},
  booktitle={Proceedings of the Neural Information Processing Systems Track on Datasets and Benchmarks},
  year={2021}
}

@inproceedings{akhtar2023exploring,
  title={Exploring the numerical reasoning capabilities of language models: A comprehensive analysis on tabular data},
  author={Akhtar, Mubashara and Shankarampeta, Abhilash and Gupta, Vivek and Patil, Arpit and Cocarascu, Oana and Simperl, Elena},
  booktitle={Findings of the Association for Computational Linguistics: EMNLP 2023},
  pages={15391--15405},
  year={2023}
}

@inproceedings{lu2023mathvista,
  title={{MathVista}: Evaluating Mathematical Reasoning of Foundation Models in Visual Contexts},
  author={Lu, Pan and Bansal, Hritik and Xia, Tony and Liu, Jiacheng and Li, Chunyuan and Hajishirzi, Hannaneh and Cheng, Hao and Chang, Kai-Wei and Galley, Michel and Gao, Jianfeng},
  booktitle={The Twelfth International Conference on Learning Representations},
  year={2024}
}

@inproceedings{zhang2024mathverse,
  title={Mathverse: Does your multi-modal llm truly see the diagrams in visual math problems?},
  author={Zhang, Renrui and Jiang, Dongzhi and Zhang, Yichi and Lin, Haokun and Guo, Ziyu and Qiu, Pengshuo and Zhou, Aojun and Lu, Pan and Chang, Kai-Wei and Qiao, Yu and others},
  booktitle={European Conference on Computer Vision},
  pages={169--186},
  year={2024},
  organization={Springer}
}

@inproceedings{wang2017deep,
  title={Deep neural solver for math word problems},
  author={Wang, Yan and Liu, Xiaojiang and Shi, Shuming},
  booktitle={Proceedings of the 2017 conference on empirical methods in natural language processing},
  pages={845--854},
  year={2017}
}

@inproceedings{he2024olympiadbench,
  title={{OlympiadBench}: A Challenging Benchmark for Promoting {AGI} with Olympiad-Level Bilingual Multimodal Scientific Problems},
  author={He, Chaoqun and Luo, Renjie and Bai, Yuzhuo and Hu, Shengding and Thai, Zhen Leng and Shen, Junhao and Hu, Jinyi and Han, Xu and Huang, Yujie and Zhang, Yuxiang and others},
  booktitle={Proceedings of the 62nd Annual Meeting of the Association for Computational Linguistics (Volume 1: Long Papers)},
  pages={3828--3850},
  year={2024}
}

@inproceedings{tsoukalas2024putnambench,
  title={{PutnamBench}: Evaluating Neural Theorem-Provers on the Putnam Mathematical Competition},
  author={Tsoukalas, George and Lee, Jasper and Jennings, John and Xin, Jimmy and Ding, Michelle and Jennings, Michael and Thakur, Amitayush and Chaudhuri, Swarat},
  booktitle={Advances in Neural Information Processing Systems},
  volume={37},
  year={2024}
}

@inproceedings{gao2024omni,
  title={{Omni-MATH}: A Universal Olympiad Level Mathematic Benchmark for Large Language Models},
  author={Gao, Bofei and Song, Feifan and Yang, Zhe and Cai, Zefan and Miao, Yibo and Dong, Qingxiu and Li, Lei and Ma, Chenghao and Chen, Liang and Xu, Runxin and others},
  booktitle={The Thirteenth International Conference on Learning Representations},
  year={2025}
}

@inproceedings{xu2022towards,
  title={Towards Robust Numerical Question Answering: Diagnosing Numerical Capabilities of {NLP} Systems},
  author={Xu, Jialiang and Zhou, Mengyu and He, Xinyi and Han, Shi and Zhang, Dongmei},
  booktitle={Proceedings of the 2022 Conference on Empirical Methods in Natural Language Processing},
  year={2022}
}

@misc{liu2023llava,
      title={Visual Instruction Tuning}, 
      author={Liu, Haotian and Li, Chunyuan and Wu, Qingyang and Lee, Yong Jae},
      publisher={NeurIPS},
      year={2023},
}

@misc{deepseekai2025deepseekr1incentivizingreasoningcapability,
      title={DeepSeek-R1: Incentivizing Reasoning Capability in LLMs via Reinforcement Learning}, 
      author={DeepSeek-AI and Daya Guo and Dejian Yang et al.},
      year={2025},
      eprint={2501.12948},
      archivePrefix={arXiv},
      primaryClass={cs.CL},
      url={https://arxiv.org/abs/2501.12948}, 
}

@article{qwen2.5,
    title   = {Qwen2.5 Technical Report}, 
    author  = {An Yang and Baosong Yang and Beichen Zhang and Binyuan Hui and Bo Zheng and Bowen Yu and Chengyuan Li and Dayiheng Liu and Fei Huang and Haoran Wei and Huan Lin and Jian Yang and Jianhong Tu and Jianwei Zhang and Jianxin Yang and Jiaxi Yang and Jingren Zhou and Junyang Lin and Kai Dang and Keming Lu and Keqin Bao and Kexin Yang and Le Yu and Mei Li and Mingfeng Xue and Pei Zhang and Qin Zhu and Rui Men and Runji Lin and Tianhao Li and Tingyu Xia and Xingzhang Ren and Xuancheng Ren and Yang Fan and Yang Su and Yichang Zhang and Yu Wan and Yuqiong Liu and Zeyu Cui and Zhenru Zhang and Zihan Qiu},
    journal = {arXiv preprint arXiv:2412.15115},
    year    = {2024}
}

@article{team2024gemma,
  title={Gemma 2: Improving open language models at a practical size},
  author={Team, Gemma and Riviere, Morgane and Pathak, Shreya and Sessa, Pier Giuseppe and Hardin, Cassidy and Bhupatiraju, Surya and Hussenot, L{\'e}onard and Mesnard, Thomas and Shahriari, Bobak and Ram{\'e}, Alexandre and others},
  journal={arXiv preprint arXiv:2408.00118},
  year={2024}
}

@article{jiang2023mistral,
  title={Mistral 7B},
  author={Jiang, Albert Q and Sablayrolles, Alexandre and Mensch, Arthur and Bamford, Chris and Chaplot, Devendra Singh and Casas, Diego de las and Bressand, Florian and Lengyel, Gianna and Lample, Guillaume and Saulnier, Lucile and others},
  journal={arXiv preprint arXiv:2310.06825},
  year={2023}
}

@misc{grattafiori2024llama3herdmodels,
      title={The Llama 3 Herd of Models}, 
      author={Aaron Grattafiori and Abhimanyu Dubey and Abhinav Jauhri et al.},
      year={2024},
      eprint={2407.21783},
      archivePrefix={arXiv},
      primaryClass={cs.AI},
      url={https://arxiv.org/abs/2407.21783}, 
}

@inproceedings{miao2020diverse,
  title={A diverse corpus for evaluating and developing English math word problem solvers},
  author={Miao, Shen-Yun and Liang, Chao-Chun and Su, Keh-Yih},
  booktitle={Proceedings of the 58th annual meeting of the Association for Computational Linguistics},
  pages={975--984},
  year={2020}
}

@inproceedings{kushman-etal-2014-learning,
    title = "Learning to Automatically Solve Algebra Word Problems",
    author = "Kushman, Nate  and
      Artzi, Yoav  and
      Zettlemoyer, Luke  and
      Barzilay, Regina",
    editor = "Toutanova, Kristina  and
      Wu, Hua",
    booktitle = "Proceedings of the 52nd Annual Meeting of the Association for Computational Linguistics (Volume 1: Long Papers)",
    month = jun,
    year = "2014",
    address = "Baltimore, Maryland",
    publisher = "Association for Computational Linguistics",
    url = "https://aclanthology.org/P14-1026/",
    doi = "10.3115/v1/P14-1026",
    pages = "271--281"
}

@inproceedings{shi-etal-2015-automatically,
    title = "Automatically Solving Number Word Problems by Semantic Parsing and Reasoning",
    author = "Shi, Shuming  and
      Wang, Yuehui  and
      Lin, Chin-Yew  and
      Liu, Xiaojiang  and
      Rui, Yong",
    editor = "M{\`a}rquez, Llu{\'i}s  and
      Callison-Burch, Chris  and
      Su, Jian",
    booktitle = "Proceedings of the 2015 Conference on Empirical Methods in Natural Language Processing",
    month = sep,
    year = "2015",
    address = "Lisbon, Portugal",
    publisher = "Association for Computational Linguistics",
    url = "https://aclanthology.org/D15-1135/",
    doi = "10.18653/v1/D15-1135",
    pages = "1132--1142"
}

@inproceedings{zhu-etal-2021-tat,
    title = "{TAT}-{QA}: A Question Answering Benchmark on a Hybrid of Tabular and Textual Content in Finance",
    author = "Zhu, Fengbin  and
      Lei, Wenqiang  and
      Huang, Youcheng  and
      Wang, Chao  and
      Zhang, Shuo  and
      Lv, Jiancheng  and
      Feng, Fuli  and
      Chua, Tat-Seng",
    editor = "Zong, Chengqing  and
      Xia, Fei  and
      Li, Wenjie  and
      Navigli, Roberto",
    booktitle = "Proceedings of the 59th Annual Meeting of the Association for Computational Linguistics and the 11th International Joint Conference on Natural Language Processing (Volume 1: Long Papers)",
    month = aug,
    year = "2021",
    address = "Online",
    publisher = "Association for Computational Linguistics",
    url = "https://aclanthology.org/2021.acl-long.254/",
    doi = "10.18653/v1/2021.acl-long.254",
    pages = "3277--3287"
}

@misc{liu2024chainofspotinteractivereasoningimproves,
      title={Chain-of-Spot: Interactive Reasoning Improves Large Vision-Language Models}, 
      author={Zuyan Liu and Yuhao Dong and Yongming Rao and Jie Zhou and Jiwen Lu},
      year={2024},
      eprint={2403.12966},
      archivePrefix={arXiv},
      primaryClass={cs.CV},
      url={https://arxiv.org/abs/2403.12966}, 
}

@article{schick2023toolformer,
  title={Toolformer: Language models can teach themselves to use tools},
  author={Schick, Timo and Dwivedi-Yu, Jane and Dess{\`\i}, Roberto and Raileanu, Roberta and Lomeli, Maria and Hambro, Eric and Zettlemoyer, Luke and Cancedda, Nicola and Scialom, Thomas},
  journal={Advances in Neural Information Processing Systems},
  volume={36},
  pages={68539--68551},
  year={2023}
}

@article{shen2023hugginggpt,
  title={Hugginggpt: Solving ai tasks with chatgpt and its friends in hugging face},
  author={Shen, Yongliang and Song, Kaitao and Tan, Xu and Li, Dongsheng and Lu, Weiming and Zhuang, Yueting},
  journal={Advances in Neural Information Processing Systems},
  volume={36},
  pages={38154--38180},
  year={2023}
}

@inproceedings{qin2023toolllm,
  title={{ToolLLM}: Facilitating Large Language Models to Master 16000+ Real-World {APIs}},
  author={Qin, Yujia and Liang, Shihao and Ye, Yining and Zhu, Kunlun and Yan, Lan and Lu, Yaxi and Lin, Yankai and Cong, Xin and Tang, Xiangru and Qian, Bill and others},
  booktitle={The Twelfth International Conference on Learning Representations},
  year={2024}
}

@article{patil2024gorilla,
  title={Gorilla: Large language model connected with massive apis},
  author={Patil, Shishir G and Zhang, Tianjun and Wang, Xin and Gonzalez, Joseph E},
  journal={Advances in Neural Information Processing Systems},
  volume={37},
  pages={126544--126565},
  year={2024}
}

@article{parisi2022talm,
  title={Talm: Tool augmented language models},
  author={Parisi, Aaron and Zhao, Yao and Fiedel, Noah},
  journal={arXiv preprint arXiv:2205.12255},
  year={2022}
}

@article{hao2023toolkengpt,
  title={Toolkengpt: Augmenting frozen language models with massive tools via tool embeddings},
  author={Hao, Shibo and Liu, Tianyang and Wang, Zhen and Hu, Zhiting},
  journal={Advances in neural information processing systems},
  volume={36},
  pages={45870--45894},
  year={2023}
}

@article{song2023restgpt,
  title={Restgpt: Connecting large language models with real-world restful apis},
  author={Song, Yifan and Xiong, Weimin and Zhu, Dawei and Wu, Wenhao and Qian, Han and Song, Mingbo and Huang, Hailiang and Li, Cheng and Wang, Ke and Yao, Rong and others},
  journal={arXiv preprint arXiv:2306.06624},
  year={2023}
}

@article{lu2023chameleon,
  title={Chameleon: Plug-and-play compositional reasoning with large language models},
  author={Lu, Pan and Peng, Baolin and Cheng, Hao and Galley, Michel and Chang, Kai-Wei and Wu, Ying Nian and Zhu, Song-Chun and Gao, Jianfeng},
  journal={Advances in Neural Information Processing Systems},
  volume={36},
  pages={43447--43478},
  year={2023}
}

@article{xu2023tool,
  title={On the tool manipulation capability of open-source large language models},
  author={Xu, Qiantong and Hong, Fenglu and Li, Bo and Hu, Changran and Chen, Zhengyu and Zhang, Jian},
  journal={arXiv preprint arXiv:2305.16504},
  year={2023}
}

@inproceedings{yuan2024easytool,
  title={{EasyTool}: Enhancing {LLM}-based Agents with Concise Tool Instruction},
  author={Yuan, Siyu and Song, Kaitao and Chen, Jiangjie and Tan, Xu and Shen, Yongliang and Kan, Ren and Li, Dongsheng and Yang, Deqing},
  booktitle={Proceedings of the 2025 Conference of the North American Chapter of the Association for Computational Linguistics: Human Language Technologies},
  year={2025}
}

@article{hsieh2023tool,
  title={Tool documentation enables zero-shot tool-usage with large language models},
  author={Hsieh, Cheng-Yu and Chen, Si-An and Li, Chun-Liang and Fujii, Yasuhisa and Ratner, Alexander and Lee, Chen-Yu and Krishna, Ranjay and Pfister, Tomas},
  journal={arXiv preprint arXiv:2308.00675},
  year={2023}
}

@article{cai2016item,
  title={Item response theory},
  author={Cai, Li and Choi, Kilchan and Hansen, Mark and Harrell, Lauren},
  journal={Annual Review of Statistics and Its Application},
  volume={3},
  number={1},
  pages={297--321},
  year={2016},
  publisher={Annual Reviews}
}

@article{luo2023wizardmath,
  title={Wizardmath: Empowering mathematical reasoning for large language models via reinforced evol-instruct},
  author={Luo, Haipeng and Sun, Qingfeng and Xu, Can and Zhao, Pu and Lou, Jianguang and Tao, Chongyang and Geng, Xiubo and Lin, Qingwei and Chen, Shifeng and Zhang, Dongmei},
  journal={arXiv preprint arXiv:2308.09583},
  year={2023}
}

@inproceedings{wang2023math,
  title={{Math-Shepherd}: Verify and Reinforce {LLMs} Step-by-step without Human Annotations},
  author={Wang, Peiyi and Li, Lei and Shao, Zhihong and Xu, RX and Dai, Damai and Li, Yifei and Chen, Deli and Wu, Yu and Sui, Zhifang},
  booktitle={Proceedings of the 62nd Annual Meeting of the Association for Computational Linguistics (Volume 1: Long Papers)},
  year={2024}
}

@inproceedings{shao2024deepseekmath,
  title={{DeepSeekMath}: Pushing the Limits of Mathematical Reasoning in Open Language Models},
  author={Shao, Zhihong and Wang, Peiyi and Zhu, Qihao and Xu, Runxin and Song, Junxiao and Bi, Xiao and Zhang, Haowei and Zhang, Mingchuan and Li, YK and Wu, Y and others},
  booktitle={The Thirteenth International Conference on Learning Representations},
  year={2025}
}

@article{schulman2020approximating,
  title={Approximating kl divergence},
  author={Schulman, John},
  journal={John Schulman’s Homepage},
  year={2020}
}

@article{goldie2025synthetic,
  title={Synthetic data generation \& multi-step rl for reasoning \& tool use},
  author={Goldie, Anna and Mirhoseini, Azalia and Zhou, Hao and Cai, Irene and Manning, Christopher D},
  journal={arXiv preprint arXiv:2504.04736},
  year={2025}
}

@article{shen2021mathbert,
  title={Mathbert: A pre-trained language model for general nlp tasks in mathematics education},
  author={Shen, Jia Tracy and Yamashita, Michiharu and Prihar, Ethan and Heffernan, Neil and Wu, Xintao and Graff, Ben and Lee, Dongwon},
  journal={arXiv preprint arXiv:2106.07340},
  year={2021}
}

@inproceedings{yang2024qwen2,
  title={{Qwen2.5-Math} Technical Report: Toward Mathematical Expert Model via Self-Improvement},
  author={Yang, An and Zhang, Beichen and Hui, Binyuan and Gao, Bofei and Yu, Bowen and Li, Chengpeng and Liu, Dayiheng and Tu, Jianhong and Zhou, Jingren and Lin, Junyang and others},
  booktitle={Advances in Neural Information Processing Systems},
  volume={37},
  year={2024}
}

\appendix

\section{Detailed Description of each Subtask}
\label{sec: description of subtasks}
In this section, we provide detailed information on each subtask in Table~\ref{tab: MWP}\~~\ref{tab: VDQ}, including 1) aspects, 2) whether it is a multimodal task, 3) size, 4) design rationale and description, and 5) all versions of the prompt we used.

\section{Experiment Settings}
\label{sec: experiment settings}
In our evaluation experiments, We conduct an evaluation of 10 representative LLMs using the PMB dataset, we adopt the model versions and sampling parameters provided in Table~\ref{tab: sampling parameters}. The version dates listed in Table~\ref{tab: sampling parameters} (e.g., GPT-4o 2024-11-20, DeepSeek-R1 2025-01-20) correspond to the API or checkpoint releases available at the time of evaluation, and reported scores should be interpreted relative to these snapshots. For experiments in Section The SOLVE module and Interactive Relative Policy Optimization, the PMB was split into training, validation, and test sets in a ratio of 8:1:1. The model was trained with the hyperparameters listed in Table~\ref{tab: IRPO param}.

\section{Validation of the Text-Answer Scoring Metric}
\label{sec: metric validation}
For open-ended text answers, responses are scored by the cosine similarity between MathBERT embeddings of the model output and the reference, with a match threshold of $0.9$. This section documents the threshold selection procedure, its agreement with human scoring, and the treatment of mathematically equivalent expressions.

\noindent\textbf{Threshold calibration.}
A random $30\%$ split of the open-ended text questions was set aside as a calibration subset, with each item independently labeled correct or incorrect by two domain experts. Candidate thresholds were swept over $[0.75, 0.95]$, and the value yielding the highest F1 against expert labels was retained. F1 differences within $[0.85, 0.92]$ are below $0.9\%$, and the relative ranking of models on the Understanding aspect is preserved across this interval, indicating that downstream conclusions are insensitive to the exact threshold choice.

\noindent\textbf{Agreement with human judgment.}
On a separately drawn validation subset, the scorer agreed with expert grading at $98.5\%$ accuracy, with a Pearson correlation of $r = 0.94$ between embedding-based and human scores. Replacing embedding scores with human scores on this subset does not change the comparative conclusions reported in Section~\ref{sec: results}.

\noindent\textbf{Mathematically equivalent expressions.}
Because embedding similarity is not a symbolic prover, a canonicalization step is applied before similarity is computed: fraction and decimal forms are unified, factored expressions are expanded, and variable orderings are standardized. A manual inspection of structurally different but mathematically equivalent reference/response pairs confirms that most such pairs exceed the $0.9$ threshold after canonicalization. Cases requiring deep symbolic identity transformations are rare and do not alter rankings on the affected subtasks.

\section{In-depth Model Analysis}
\label{sec: model analysis}
\begin{figure}[hbtp]
    \centering
    \includegraphics[width=\linewidth]{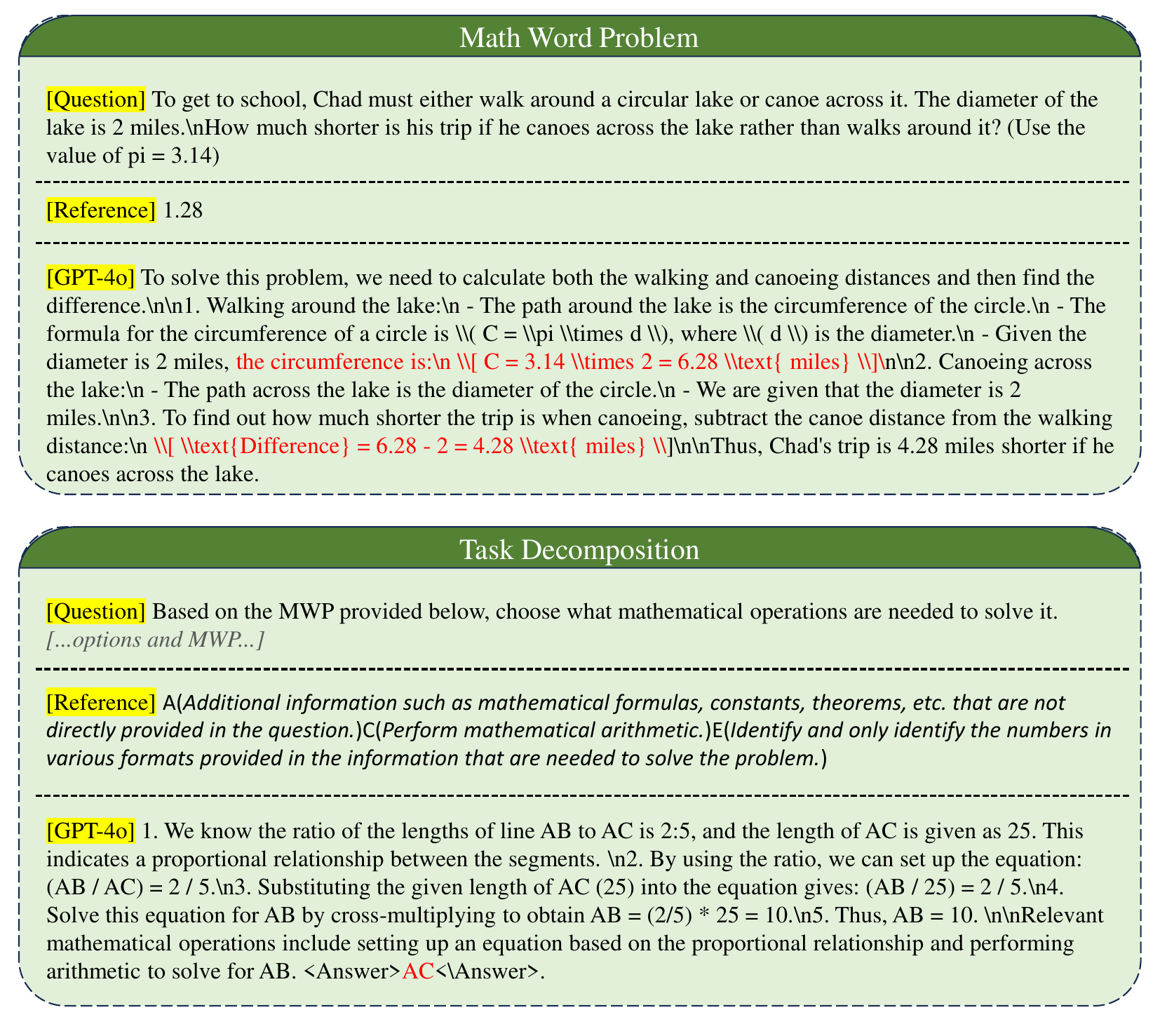}
    \caption{An example illustrating GPT-4o failed to answer an MWP question, and this can be oriented to overlook the significance of numerical information identification in its  TD subtask.}
    \label{fig: cs1}
\end{figure}

PyraMathBench (PMB), which builds upon Piaget’s cognitive theory, offers a more comprehensive evaluation framework by organizing mathematical tasks into a pyramid of four capability levels, specifically designed to address and evaluate these deficiencies by rigorously assessing LLMs' performance on both foundational and advanced tasks. Introducing the subtask decomposition mechanism in PMB enhances its ability to pinpoint logical reasoning errors with precision. For example, in Figure~\ref{fig: cs1}, GPT-4-o failed to answer an MWP. Upon reviewing its response in the corresponding Task Decomposition subtask, it became apparent that GPT-4-o overlooked the significance of numerical information identification. A closer examination of its reasoning process revealed that the error stemmed from a misinterpretation of the relationship between the diameter and the distance, leading to an information extraction error, so this error could be attributed to the wrong identification of information. This case demonstrates PMB’s effectiveness in detecting inference errors or omissions.

\begin{figure}[hbtp]
    \centering
    \includegraphics[width=\linewidth]{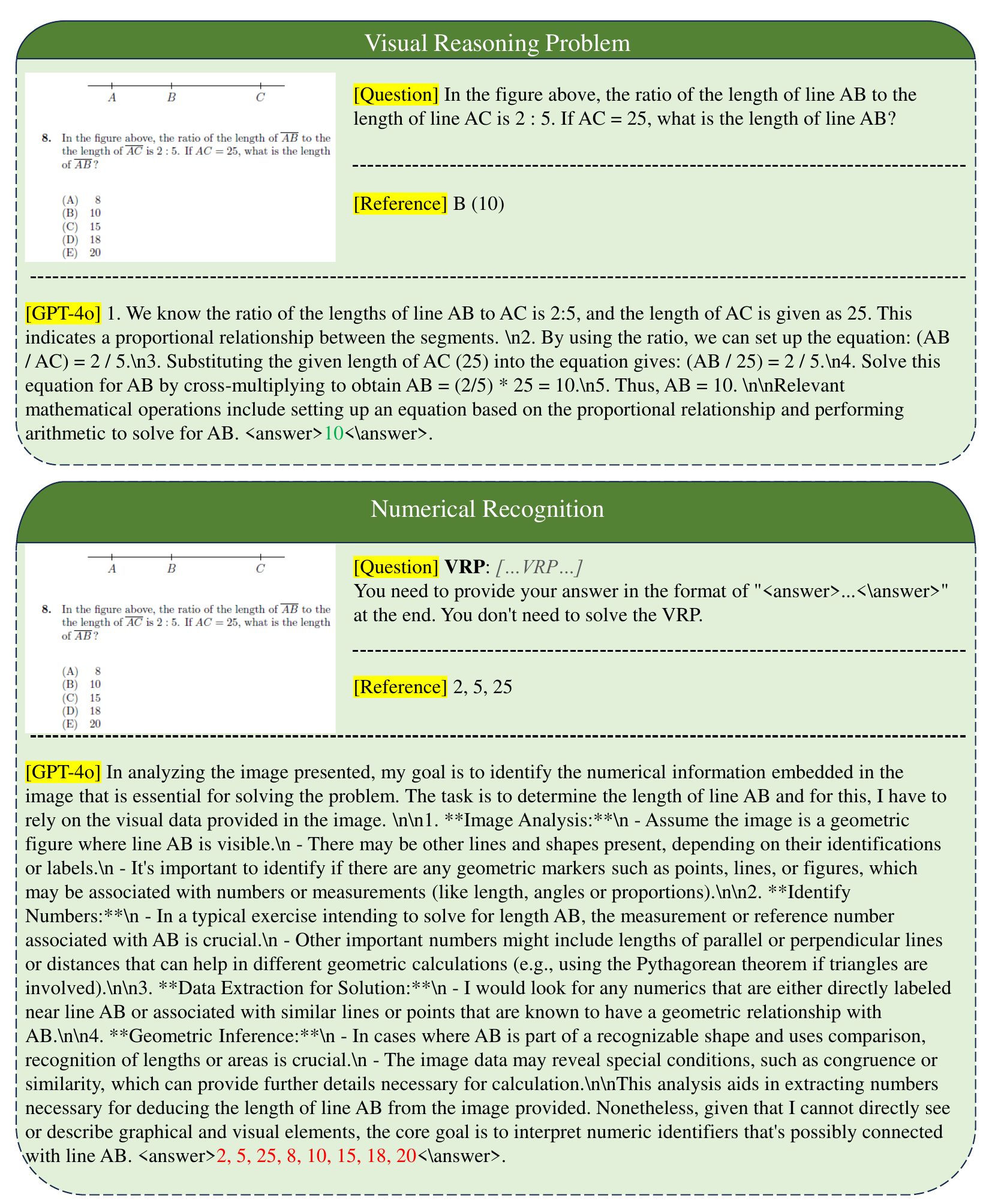}
    \caption{An example illustrating GPT-4o mainly relies on text information to solve multimodal problems instead of visual information.}
    \label{fig: cs2}
\end{figure}

In Figure~\ref{fig: cs2}, GPT-4o appears to solve a Visual Reasoning Problem from MathVista correctly. However, a closer inspection of the corresponding Numerical Recognition subtask reveals that GPT-4o's solution is based entirely on textual information, with no contribution from the visual elements of the problem. Existing multimodal datasets often lead to overestimating MLLMs’ visual reasoning abilities, as humans may incorrectly assume that the model is leveraging visual cues. PMB addresses this gap by quantifying the extent to which MLLMs rely on various modalities by segmenting the task into separate multimodal subtasks. This approach allows PMB to effectively identify and highlight the erroneous over-reliance on text, thereby providing a more rigorous assessment of the model’s true multimodal capabilities.

\begin{figure}[hbtp]
    \centering
    \includegraphics[width=\linewidth]{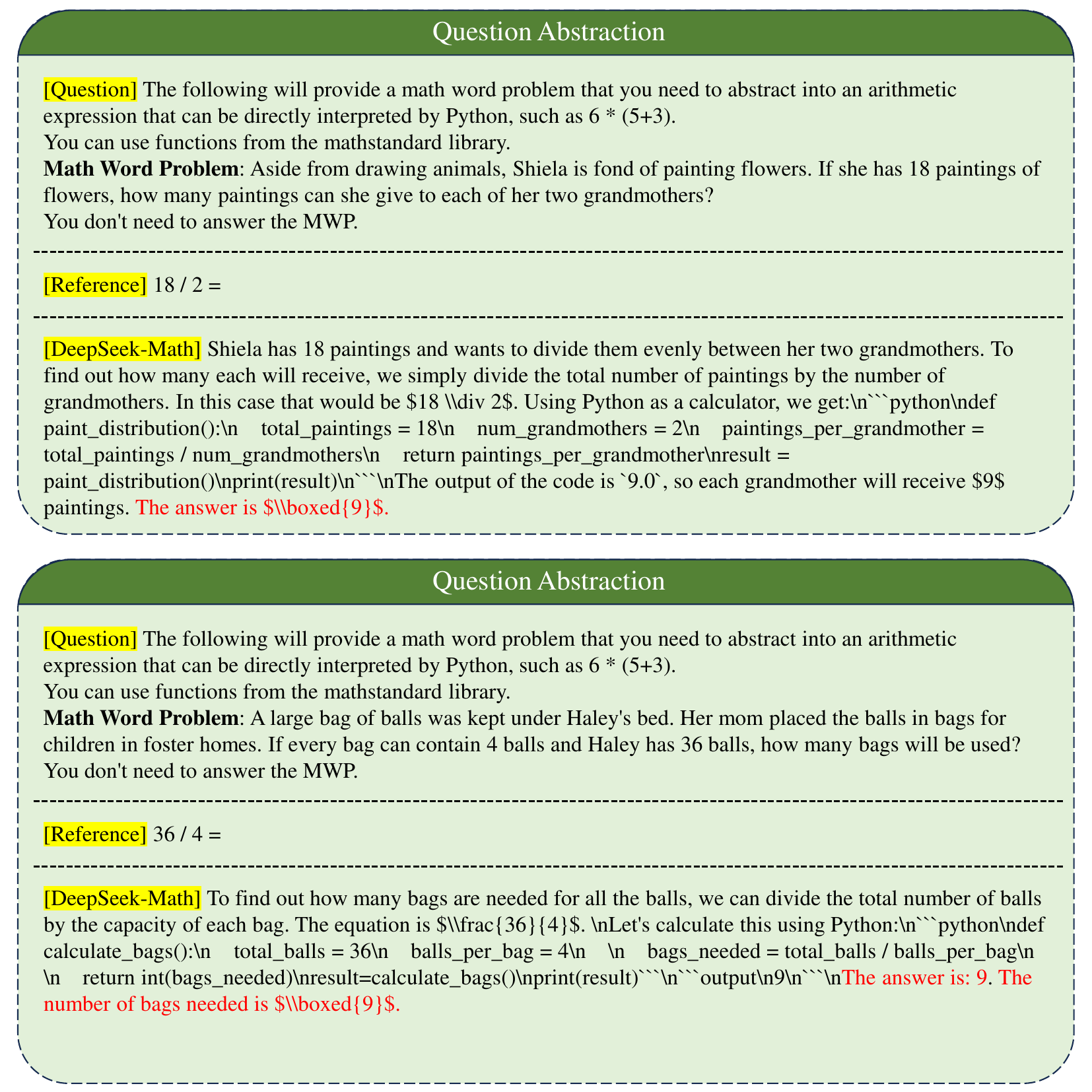}
    \caption{Two examples illustrating DeepSeek-Math failed to answer QA questions due to inadequate instruction following ability, and its answers are given in different formats.}
    \label{fig: cs3}
\end{figure}

For LLMs that were fine-tuned on mathematical tasks like DeepSeek-Math and Qwen2.5-Math, inadequate instruction following ability is one of the reasons for their poor performance on multiple subtasks. Additionally, DeepSeek-Math's inconsistent answer formats contributed to its lower performance. In Figure~\ref{fig: cs3}, DeepSeek-Math ignores prompts asking to abstract the question into an expression and doesn't answer the MWP, instead directly calculating answers. And in the two QA questions, DeepSeek-Math responded in different formats, neither of which matches the indication of the prompt.

\begin{figure}[hbtp]
    \centering
    \includegraphics[width=\linewidth]{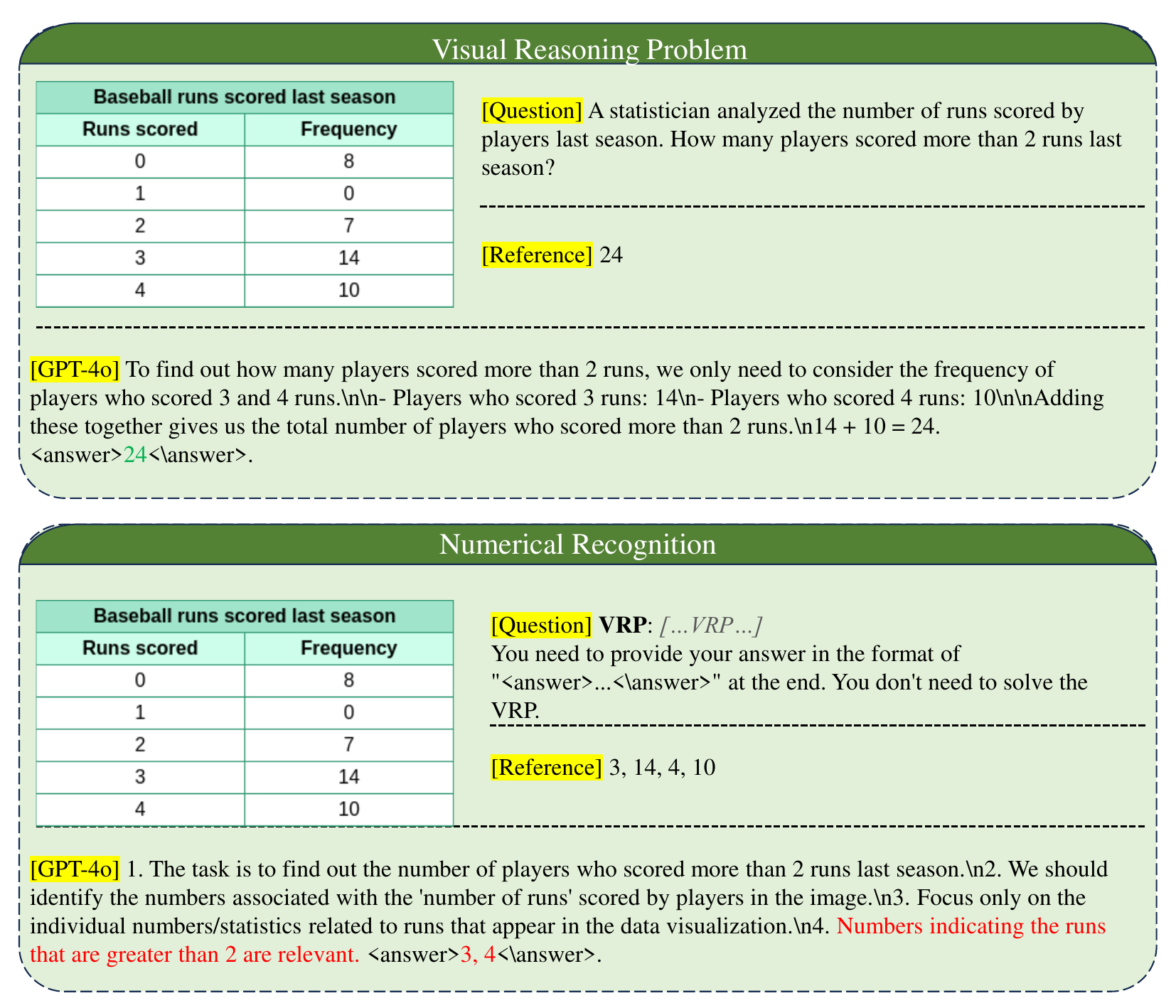}
    \caption{An example illustrating GPT-4o failed to extract the target numbers from the visual table but solved the corresponding visual reasoning problem correctly through logical reasoning.}
    \label{fig: cs4}
\end{figure}

Through the case analysis in Figure~\ref{fig: cs4}, we identified that the failure of MLLMs in NR tasks stems primarily from their inability to extract only the required numbers. Although GPT-4o recognizes numbers with relatively high accuracy, the numbers in the answer are randomly selected from the image without focusing on the relevant areas necessary for solving the problem. GPT-4o's response to the corresponding visual reasoning problem revealed that while LLMs struggle to identify relevant data in lower-level tasks, they effectively discard incorrect answers through logical reasoning in higher-level tasks, leading to a lesser performance degradation in those cases.

In VDQ tasks, MLLMs exhibit prominent hallucinations, resulting in inferences and analyses that deviate from the actual content of the image. In OC tasks, MLLMs fail not only due to their inability to select the correct objects based on instructions but also due to poor performance in counting large, patterned groups. Hence, MLLMs struggle to extract meaningful information from images when addressing visual reasoning problems, relying primarily on text-based data. This suggests that some previous work~\cite{liu2024chainofspotinteractivereasoningimproves} focused on enhancing feature extraction through key region-of-interest identification in images may fail to yield sufficiently satisfactory results in mathematical contexts.

\begin{figure}[hbtp]
    \centering
    \includegraphics[width=\linewidth]{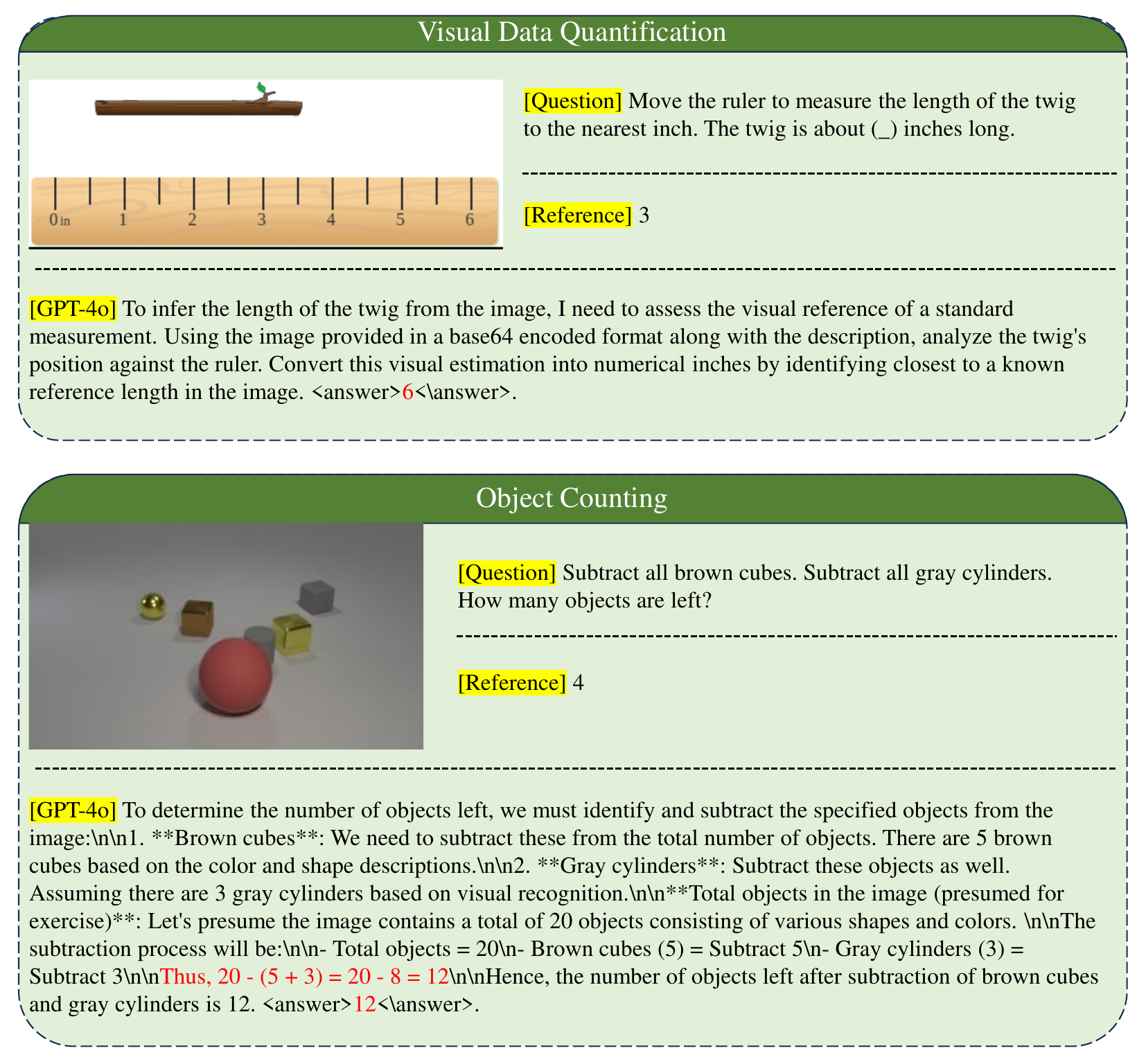}
    \caption{Two examples illustrating GPT-4o failed to answer a VDQ question and an OC question.}
    \label{fig: cs5}
\end{figure}

\begin{table*}[htbp] 
 \centering
 \renewcommand\arraystretch{1}
 \begin{tabular}{p{1in}p{4in}}
 \toprule
 \textbf{Model} & \textbf{Hyperparameters}\\
 \midrule
 GPT-4o & model=GPT-4O 2024-11-20 version, temperature=0.7, top\_p=0.9, max\_tokens=1000 \\
 \midrule
 GPT-4o-mini & model=GPT-4O-mini 2024-7-18 version, temperature=0.7, top\_p=0.9, max\_tokens=1000\\
 \midrule
 LLaVA & model=LLava-v1.5-7B, temperature=0.7, top\_p=0.9, max\_tokens=1000, quantization=Q4\_0\\
 DeepSeek-R1 & model=DeepSeek-R1 2025-1-20 version, temperature=0.7, top\_p=0.9, max\_tokens=1000\\
 \midrule
 DeepSeek-MATH & model=DeepSeek-Math-7B-rl, temperature=0.7, top\_p=0.9, max\_tokens=1000, quantization=Q8\_0\\
 \midrule
 Qwen & model=Qwen2.5-14B-Instruct, temperature=0.7, top\_p=0.9, max\_tokens=1000, quantization=Q8\_0\\
 \midrule
 Qwen-MATH & model=Qwen2.5-Math-7B-Instruct, temperature=0.7, top\_p=0.9, max\_tokens=1000, quantization=Q8\_0\\
 \midrule
 Llama3.1 & model=Llama3.1-8B-Instruct, temperature=0.7, top\_p=0.9, max\_tokens=1000, quantization=Q8\_0\\
 \midrule
 Gemma2 & model=Gemma2-9B-Instruct, temperature=0.7, top\_p=0.9, max\_tokens=1000, quantization=Q8\_0\\
 \midrule
 Mistral & model=Mistral-9B-Instruct, temperature=0.7, top\_p=0.9, max\_tokens=1000, quantization=Q8\_0\\
 \bottomrule
 \end{tabular}
 \caption{Hyperparameter for LLMs and MLLMs in main experiments.}
 \label{tab: sampling parameters}
 \end{table*}

\section{Design of SOLVE Module}
\label{sec: design of solve module}
\subsection{Functions of the SOLVE Module}
The SOLVE supports several functions, including 1) Arithmetic, 2) Equation Solving, 3) Sorting, and 4) Knowledge Explanation.
Here are the detailed descriptions of each function:
\begin{itemize}[topsep=0pt,itemsep=0pt,parsep=0pt,partopsep=0pt]
    \item \textbf{Arithmetic.} Used to calculate expressions, input a math expression in python expression format or latex format, return the corresponding result, with 6 decimal places retained.
    \item \textbf{Equation Solving.} A equation solver. Input one or more unknowns and their corresponding number of equations, return the solution to this equation or system of equations, with 6 decimal places retained.
    \item \textbf{Sorting.} Sorts a set of numbers, enter a sequence of numbers and arrange them in ascending or descending order as required.
    \item \textbf{Knowledge Explanation.} Supplies mathematical knowledge (e.g., formulas, definitions, and theorem proofs) in response to LLM queries from vector database.
\end{itemize}
\subsection{Algorithm for Dynamic Assessment of LLM Ability}
SOLVE employs a rule-based mechanism to estimate the difficulty and discrimination of questions, and adaptively adjusting their estimated ability level based on performance across questions of varying difficulty. The algorithm is grounded in psychometric principles, drawing inspiration from adaptive testing frameworks and item response theory (IRT). Key design principles include: 1) \textbf{Adaptive Updates}: The LLM’s ability estimate (denoted as \(\theta\)) is adjusted iteratively based on its performance, with larger updates for unexpected outcomes (e.g., solving a far more difficult problem or failing an easier one). 2) \textbf{Difficulty-Aware Feedback}: Each problem’s discrimination \(\alpha\) and difficulty \(\beta\) is predefined via rule-based methods depicted in Subsection~\ref{sec: diff and dis}, enabling systematic comparisons between (\(\theta\)), \(\alpha\), and \(\beta\)) using the equation \(P(X=1|\theta)=\frac{1}{1+e^{-\alpha(\theta-\beta)}}\), discrimination controls the sensitivity of the expectation to the \(\theta-\beta\) difference. 3) \textbf{Nonlinear Response}: Updates are scaled according to the discrepancy between (\(\theta\)) and \(\beta\)) and the accuracy of LLM response, ensuring proportional adjustments—significant for surprising results, minor for expected ones. Specifically, the ability estimate \(\theta\) is updated using a prediction error-driven adjustment: \(\theta_{t+1}=\theta + \lambda \cdot (\theta_t - \beta)\), Here, \(\lambda\) is a learning rate modulating the step size, we adopt 0.01 in experiments. Steps 2–4 repeat across multiple problems, progressively refining \(\theta\) to converge toward the LLM’s true ability level.

\subsection{Estimation of Question Difficulty and Discrimination}
\label{sec: diff and dis}
The estimation of difficulty and discrimination of mathematical expressions is designed to quantify computational complexity through a multi-dimensional evaluation framework. This methodology combines structural analysis of the expression syntax with semantic evaluation of computational outcomes, implemented through three primary assessment dimensions:
\begin{itemize}
    \item \textbf{Operator and Function Complexity.} Arithmetic operations receive lowest weights. Common mathematical functions are assigned customized weights reflecting their computational demands. Advanced operations or functions (e.g., calculus) are assigned with highest weights.
    \item \textbf{Structural Complexity.} Expressions with multiple distinct variables and higher recursive depth are assigned with higher scores. 
    \item \textbf{Result-Type Evaluation.} Scores increase along the complexity spectrum: Integer < Rational < Floating-point < Irrational.
\end{itemize}
The coefficients assigned to each assessment dimension depend on the specific task and actual usage scenario. In this work, we extracted 1000 math word problems from PMB and initialized the coefficients based on the accuracy of each LLMs.

\section{Detailed Implementation of IRPO}
\label{sec: implementation of IRPO}
IRPO is a new RL framework that extends GRPO by incorporating tool usage. In this section, we explain the implementation of IRPO reward functions.

For a given question q, the IRPO simultaneously samples two sets of outputs, one from the base policy model \(\pi_{\theta}\), and another from the tool-augmented policy model \(\pi^{\tau}_{\theta}\), which include both intermediate tool-related responses and final LLM responses. Therefore, there are two kinds of reward functions, one is to score the response that provides the final answer, and the other is to score the tool call request of the large model. We have explained the first kind of reward function in main content. For the second kind of reward function (i.e., Equation 4), the reward function is structured to address three critical challenges in guiding LLMs to appropriately invoke external tools for mathematical problem-solving: (1) incentivizing tool usage only when necessary, (2) discouraging redundant tool calls when the model can independently solve subproblems, and (3) balancing tool reliance with the LLM’s inherent proficiency.

The first term, \(\mathbb I_{q_{sub}}\left(q_{ij}\right)\times \sqrt{1-\text{Acc}^2\left(q_{ij}\right)}\), penalizes unnecessary tool calls based on LLM's performance on subtasks within the PMB. Here, \(\text{Acc}\left(q_{ij}\right)\) measures the LLM’s standalone accuracy on \(q_{ij}\) subtask , precomputed from a diagnostic evaluation phase. The square root function ensures the penalty scales non-linearly: as \(\text{Acc}\left(q_{ij}\right)\) increases (indicating strong intrinsic capability), the reward decreases sharply, discouraging redundant tool usage.

The second term, \(\frac{e^{-\text{Mean}(r_i)}-e^{-1}}{1-e^{-1}}\), introduces a dynamic baseline derived from the average reward of the default policy. During training, this term normalizes rewards to the range [0,1] using exponential smoothing. A high \(\text{Mean}(r_i)\) (indicating competent standalone performance) reduces the reward for tool invocation, steering the model toward self-reliance. Conversely, low baseline rewards amplify incentives for tool calls.

The third term, \(\frac{1}{1+e^{-a_q(\theta-b_q)}}\), employs a logistic function to model the likelihood of the LLM (with proficiency \(\theta\)) generating correct answers without tools. Parameters (discrimination) and (difficulty) are task-specific, calibrated via Item Response Theory. This term ensures the reward accounts for the LLM’s evolving skill level: as \(\theta\) surpasses \(\beta\), the reward for tool-free solutions increases, promoting autonomous problem-solving aligned for LLMs with high proficiency.

\section{Data Construction Process}
\label{sec: data construction appendix}
\subsection{Human Annotation Detail}
The dataset annotation is conducted by six experts proficient in high school-level mathematics. The subtask questions are evenly distributed among the six experts for annotation, while the corresponding answers require validation by at least two experts. Table~\ref{tab: annotation direct} and ~\ref{tab: MLLM annotation direct} provides specific annotation guidance for each expert.

To ensure annotation consistency, we measured inter-annotator agreement using macro-averaged Cohen's Kappa, obtaining a substantial agreement score of $\kappa=0.86$. For disagreement resolution, any unresolved discrepancies between the two validating experts were escalated to a third expert annotator who served as an impartial arbitrator to determine the final annotation. This multi-stage validation ensures high-quality and consistent subtask decomposition across the entire dataset.

\onecolumn
\begin{longtable}{|p{5in}|}
 \toprule
 \hspace{0.5cm}This document provides a comprehensive guide for annotating Math Word Problems (MWPs) into the 9 specified subtasks. These guidelines are designed to assist experts in evaluating the mathematical complexity of MWPs and determining how each of the 9 subtasks is incorporated or not in the problem. Experts should follow this process step-by-step to ensure a consistent and thorough annotation of each problem. You can refer to the following example question and the example annotations in each subtesks.\\
 \\
 \textbf{Example Question}: Tom has a rectangular garden with an area of 30 square meters and a perimeter of 22 meters. What are the dimensions of the garden?
 \\
 1. Question Abstraction\\
Goal: To convert the problem into a structured mathematical representation.\\
Steps to Annotate:\\
Check for mathematical operations or relationships: Look for direct operations or relationships that can be mathematically represented (e.g., addition, multiplication, algebraic expressions).\\
Identify variables or unknowns: Look for words that indicate variables or unknowns that need to be solved for (e.g., “x”, “y”, “total”, “amount”).\\
Extract mathematical expressions: If there are relationships in the problem, translate them into equations, arithmetic expressions, or sorting list.\\
Example Annotation: \\
Unknown variables: L, W. Equations: \(2 \times (L+W)=22\), \(L \times W=30\)\\
\\
2. Task Decomposition\\
Goal: To break down the problem into steps for solving.\\
Steps to Annotate:\\
Identify the main goal: What is the problem asking for (e.g., finding an unknown, calculating a total)?\\
Determine substeps: Select the sub steps required from A-I below:\\
A: Additional information such as mathematical formulas, constants, theorems, etc. that are not directly provided in the question.\\
B: Solve an equation or system of equations.\\
C: Perform mathematical arithmetic.\\
D: Sort or compare the data in the question.\\
E: Identify and only identify the numbers in various formats provided in the information that are needed to solve the problem.\\
F: Identify the numerical unit(s) required to obtain the answer\\.
G: Identify and only identify the numbers in various formats provided in the image that are needed to solve the problem.\\
H: Quantify data in images that are not directly presented in numerical terms.\\
I: Count the number of certain objects in the picture.\\
Suggest completing it at the end.\\
Example Annotation:\\
ABCDEFGHI\\
\\
3. Math Knowledge\\
Goal: To evaluate whether the model needs to apply advanced mathematical concepts not explicitly stated in the problem.\\
Steps to Annotate:\\
Check for implicit knowledge: Look for problems that require knowledge of constants, special numbers (like pi, e), or advanced mathematical formulas (e.g., quadratic formula, trigonometric identities).\\
Select formula: Select the index of the desired formula or constant from the formula library.\\
Example Annotation:\\
5(Quadratic Formula)\\
\\
4. Arithmetic\\
Goal: To evaluate the basic arithmetic operations ability of LLMs.\\
Steps to Annotate:\\
Check for simple operations: Identify basic arithmetic operations such as addition, subtraction, multiplication, division, exponentiation, or square roots.\\
Check for complex arithmetic expressions: Some problems may involve multiple operations that need to be solved step by step.\\
Example Annotation:\\
Arithmetic: \(\frac{-(-11)\pm \sqrt{(-11)^2-4\times30}}{2}\)\\
Answer: 6\\
\\
5. Equation Solving\\
Goal: To solve equations involving one or more variables.\\
Steps to Annotate:\\
Identify equations: Look for sentences that imply an equation that can be solved for an unknown (e.g., "x + 5 = 10").\\
Identify types of equations: Distinguish between linear, quadratic, or higher-degree equations, as well as systems of equations.\\
Equation abstraction: Extract the unknown variables and corresponding equations from the problem, and it is recommended to use xyz as the unknown variable name.\\
Example Annotation:\\
Unknown variables: L, W. Equations: \(2 \times (L+W)=22\), \(L \times W=30\)\\
Answer: \(L=6\), \(W=5\)\\
\\
6. Sorting
Goal: To arrange numbers or objects in a specific order.\\
Steps to Annotate:\\
Identify ordering criteria: Look for instructions that require the arrangement of numbers(ascending/descending), objects, or values based on a given condition.\\
Check for number: Extract the numbers that need to be sorted from the question and use the sorted result as the answer.
Example Annotation:\\
Sort: 6, 5, descending(Sort the solution results of the equation and determine the specific values of L and W.)
Answer: 6, 5\\
\\
7. Number Conversion\\
Goal: To convert numbers between formats.\\
Steps to Annotate:\\
Identify number types: Check for numbers expressed in various forms (e.g., words, scientific notation, fractions) and convert to Arabic numerals.\\
Check for required numbers: Identify which numbers are truly needed to solve the problem and eliminate irrelevant numbers.\\
Example Annotation:\\
30, 22\\
\\
8. Unit Conversion
Goal: To convert between different units of measurement.\\
Steps to Annotate:\\
Identify units: Look for units mentioned in the problem (e.g., meters, kilograms, degrees Celsius).\\
Check for necessary conversions: Determine if the problem requires converting between units (e.g., from kilometers to miles, Celsius to Fahrenheit).\\
Apply conversion factors: Use known conversion rates or formulas to convert units.\\
Example Annotation:\\
30 (square meters), 22 (meters)\\
\\
\textbf{Final Notes}:\\
Cross-check each problem: Not all problems will involve every subtask. Some MWPs may require only one or a few subtasks, while others may require a combination.\\
Record only relevant subtasks: If a subtask is not applicable to the problem, do not annotate it. Focus on what is essential for solving the problem.\\
Be Objective: Extract only the data that is relevant to the specific subtask. Avoid over-annotating or adding interpretations that go beyond the task requirements.\\
 \bottomrule
 \caption{Guidance for annotate 9 text-only subtasks.}
 \label{tab: annotation direct}
\end{longtable}

\begin{longtable}{|p{5in}|}
 \toprule
 \hspace{0.5cm}This document provides a comprehensive guide for annotating Math Word Problems (MWPs) into the 9 specified subtasks. These guidelines are designed to assist experts in evaluating the mathematical complexity of MWPs and determining how each of the 9 subtasks is incorporated or not in the problem. Experts should follow this process step-by-step to ensure a consistent and thorough annotation of each problem. You can refer to the following example question and the example annotations in each subtesks.\\
 \\
 \textbf{Example Question}: What is the result of dividing the sum of the times pointed to by the minute hands of all the clocks in the picture by two?\\
 \textbf{Example Image}:\\
\begin{minipage}{0.7\textwidth}
\centering
\includegraphics[width=3in, height=2.7in]{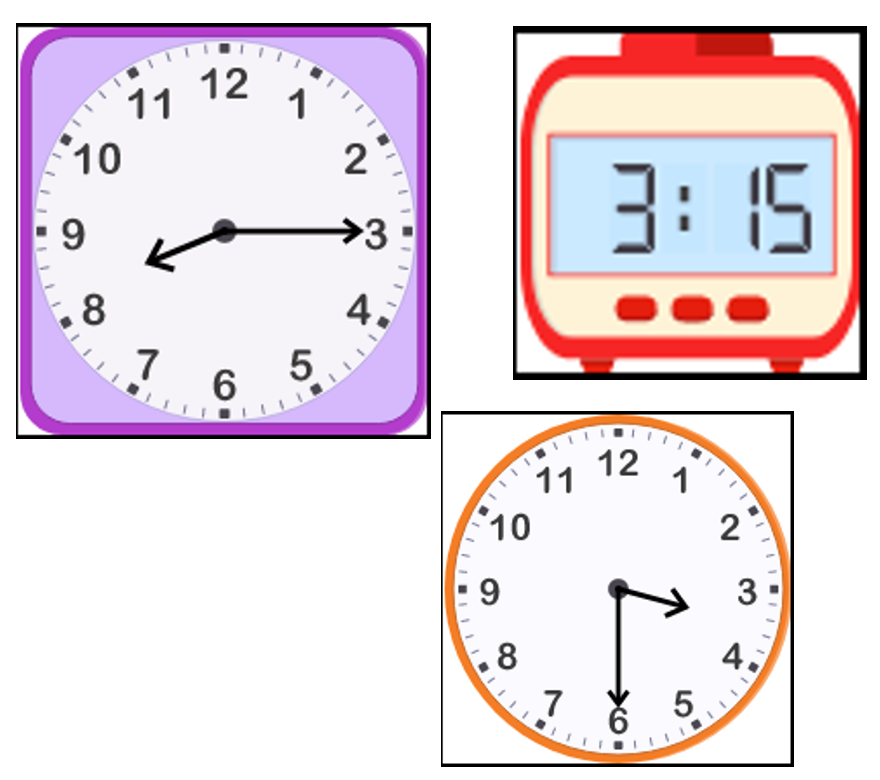}
\end{minipage}\\
\\
1. Numeral Recognition\\
Goal: Look for visual content in the image that contains numbers, symbols, or mathematical expressions. (e.g.Numbers, Variables/Constants, Formulas or Expressions)\\
Steps to Annotate:\\
Extracting Numbers and Symbols: Carefully select all visible numerical elements, ensuring that only those that are directly relevant to the problem are included. For example, in a mathematical formula within an image, extract the numbers and symbols that make up the equation.\\
Exclude Invalid Information: If the image contains irrelevant or distracting elements that might confuse the LLM, such as extra digits or misplaced symbols.\\
Example Annotation:\\
15(minutes of the digital clock, 3 shouldn't be annotated since it is not used for solving the problem.)\\
\\
2. Visual Data Quantification\\
Goal: Look for elements in the image that involve visual representations of data that aren’t explicitly presented as numbers.\\
Steps to Annotate:\\
Check for quantified data: Check if the image contain any visual data (like clocks, rulers, or diagrams) that requires interpretation and transformation into a numerical value.\\
Extract useful data: Mark the quantified numbers required for problem-solving from the image and record their corresponding labels.\\
Example Annotation:\\
Minutes of the left top analog clock: 15\\
Minutes of the bottom analog clock: 30\\
\\
3. Object Counting\\
Goal: Look for descriptions or instructions that request a specific count of objects or items within the image. These objects can vary from physical items to abstract representations.\\
Steps to Annotate:\\
Identify and Isolate Objects: Find all objects that match the description and isolate them visually. If objects are grouped or clustered together, break down the grouping to count each one individually.\\
Handle Ambiguities: If objects are obscured or difficult to distinguish, note the possible error margins or uncertainties, ensuring clear documentation of assumptions.\\
Count the Objects: Count each distinct object and record the total. Be mindful of specific instructions that may request a subset of objects (e.g., only counting certain colors or shapes).\\
Example Annotation:\\
How many clocks are there in the picture?: 3\\
 \bottomrule
 \caption{Guidance for annotate 3 multi-modal subtasks.}
 \label{tab: MLLM annotation direct}
\end{longtable}

\twocolumn
\subsection{Visualization of Tabular Data}
The Visual Reasoning Problems subtask of PMB includes table tasks visualized using table data. The conversion process works by first preparing and structuring the table data, then visually rendering it into an image format using a variety of stylistic choices. The detailed processes are as follows:
\begin{itemize}
    \item \textbf{Initial Table Processing}: The process begins by determining the maximum width and overall height of the columns within the table, which is essential for correctly formatting the table when converted into an image. Then we initiate the metadata of the table, including width, height, labels, index, and caption.
    \item \textbf{Styling Selection}: The appearance of the table image is controlled by various randomly assigned style parameters. This includes decisions on the color and style of the divider lines between columns and rows, as well as the background colors for the cells. For example, the divider lines can be styled with different patterns (solid, dashed, dotted, etc.), and their color is either black or a randomly generated similar color depending on the selected style.
    \item \textbf{Column Definitions and Layout}: Then we defines how each column of the table should be displayed, including the alignment of text (left-aligned for the index column). The column widths are adjusted based on the length of the content in each column, ensuring that the text does not overflow or become misaligned when rendered in the image. The table is then defined using the Table object, which includes all of these layout and stylistic parameters.
    \item \textbf{Table Rendering}: Using the defined table, we proceeds to render it into an image This step involves integrating the column definitions, coloring, and divider styles into the visual representation of the table. The final table image is generated with a specified size and saved as a PNG file, named according to the question index.
\end{itemize}

\begin{table}[ht]
\centering
\begin{tabular}{ll}
\hline
\textbf{Parameter} & \textbf{Value} \\
\hline
Learning rate & 5e-6 \\
Adam $\beta_1$ & 0.9 \\
Adam $\beta_2$ & 0.99 \\
Weight decay & 0.1 \\
Warmup ratio & 0.1 \\
LR scheduler & Cosine \\
Train batch size (per device) & 1 \\
Gradient accumulation steps & 4 \\
Number of generations & 4 \\
Number of generations with tools & 4 \\
Max prompt length & 256 \\
Max completion length & 1024 \\
Number of training epochs & 2 \\
Max gradient norm & 0.1 \\
Temperature & 0.7 \\
Top-k & 10 \\
Top-p & 0.9 \\
\hline
\end{tabular}
\caption{Training hyperparameters used for experiments in Section The SOLVE module and Interactive Relative Policy Optimization.}
\label{tab: IRPO param}
\end{table}

\begin{table*}[ht]
\centering
\begin{tabular}{ccccc}
\hline
\multicolumn{2}{c|}{\textbf{Subtask}} & \multicolumn{1}{c|} {\textbf{Aspect}} & \multicolumn{1}{c|}{\textbf{Multi-modal}} & \textbf{Size} \\ 
\hline
\multicolumn{2}{c|}{Math Word Problem (MWP)} & \multicolumn{1}{c|}{Complex Reasoning} & \multicolumn{1}{c|}{No} & 5476\\ 
\hline\hline
\multicolumn{5}{c}{\textbf{Description}}\\ 
\hline
\multicolumn{5}{p{5in}}{\hspace{0.5cm}Assesses the model’s ability to solve mathematical problems presented in natural language and reasoning through complex, real-world problems and translating them into mathematical solutions.}\\ 
\hline\hline
\multicolumn{5}{c}{\textbf{Prompt}}\\ 
\hline
\multicolumn{3}{c|}{System} & \multicolumn{2}{c}{Human}\\
\hline
\multicolumn{3}{p{3.3in}|}{\hspace{0.5cm}You are a helpful AI robot, you can solve mathematical problem accurately.\textbackslash nThink step by step and answer the following math word problem.} & \multicolumn{2}{p{1.7in}}{**Question**: \textbackslash n\{question\}\textbackslash n\textbackslash nYou need to provide your answer in the format of "<answer>...<\textbackslash answer>" at the end. If the result is a floating-point number, round to six decimal places.}\\
\hline
\end{tabular}
\caption{Detailed description of subtask Math Word Problem.}
\label{tab: MWP}
\end{table*}

\begin{table*}[htbp]
\centering
\begin{tabular}{ccccc}
\hline
\multicolumn{2}{c|}{\textbf{Subtask}} & \multicolumn{1}{c|} {\textbf{Aspect}} & \multicolumn{1}{c|}{\textbf{Multi-modal}} & \textbf{Size} \\ 
\hline
\multicolumn{2}{c|}{Object Counting (OC)} & \multicolumn{1}{c|}{Numerical Parsing} & \multicolumn{1}{c|}{Yes} & 107\\ 
\hline\hline
\multicolumn{5}{c}{\textbf{Description}}\\ 
\hline
\multicolumn{5}{p{5in}}{\hspace{0.5cm}This subtask requires models to count specified objects in an image based on a given description. It tests the models' visual reasoning and object recognition skills.}\\ 
\hline\hline
\multicolumn{5}{c}{\textbf{Prompt}}\\ 
\hline
\multicolumn{3}{c|}{System} & \multicolumn{2}{c}{Human}\\
\hline
\multicolumn{3}{p{3.3in}|}{\hspace{0.5cm}You are a helpful AI robot, you can solve mathematical problem accurately.\textbackslash nIdentify the number of specified objects from the following image.} & \multicolumn{2}{p{1.7in}}{**Target Object**: \textbackslash n\{question\}\textbackslash n\textbackslash nYou need to provide your answer in the format of "<answer>...<\textbackslash answer>" at the end. You don't need to solve the VRP.}\\
\hline
\end{tabular}
\caption{Detailed description of subtask Object Counting.}
\label{tab: OC}
\end{table*}

\begin{table*}[htbp]
\centering
\begin{tabular}{ccccc}
\hline
\multicolumn{2}{c|}{\textbf{Subtask}} & \multicolumn{1}{c|} {\textbf{Aspect}} & \multicolumn{1}{c|}{\textbf{Multi-modal}} & \textbf{Size} \\ 
\hline
\multicolumn{2}{c|}{Visual Reasoning Problem (VRP)} & \multicolumn{1}{c|}{Complex Reasoning} & \multicolumn{1}{c|}{Yes} & 1928\\ 
\hline\hline
\multicolumn{5}{c}{\textbf{Description}}\\ 
\hline
\multicolumn{5}{p{5in}}{\hspace{0.5cm}This subtask evaluates the model's ability to combine textual and visual information for solving mathematical problems. MLLMs need to reason across multiple modalities and extract relevant insights from both text and images. The tasks include various types such as geometry problems, VQA, and statistic reasoning.}\\ 
\hline\hline
\multicolumn{5}{c}{\textbf{Prompt}}\\ 
\hline
\multicolumn{3}{c|}{System} & \multicolumn{2}{c}{Human}\\
\hline
\multicolumn{3}{p{3.3in}|}{\hspace{0.5cm}You are a helpful AI robot, you can solve mathematical problem accurately.\textbackslash nThink step by step and answer the following visual reasoning problem based on the following image.} & \multicolumn{2}{p{1.7in}}{**Question**: \textbackslash n\{question\}\textbackslash n\textbackslash nYou need to provide your answer in the format of "<answer>...<\textbackslash answer>" at the end. If the result is a floating-point number, round to six decimal places.}\\
\hline
\end{tabular}
\caption{Detailed description of subtask Visual Reasoning Problem.}
\end{table*}

\begin{table*}[htbp]
\centering
\begin{tabular}{ccccc}
\hline
\multicolumn{2}{c|}{\textbf{Subtask}} & \multicolumn{1}{c|} {\textbf{Aspect}} & \multicolumn{1}{c|}{\textbf{Multi-modal}} & \textbf{Size} \\ 
\hline
\multicolumn{2}{c|}{Question Abstraction (QA)} & \multicolumn{1}{c|}{Understanding} & \multicolumn{1}{c|}{No} & 8883\\ 
\hline\hline
\multicolumn{5}{c}{\textbf{Description}}\\ 
\hline
\multicolumn{5}{p{5in}}{\hspace{0.5cm}This subtask requires LLMs to convert natural language problems into solvable structured mathematical representations, including arithmetic, equations, and sorting numbers.}\\ 
\hline\hline
\multicolumn{5}{c}{\textbf{Prompt}}\\ 
\hline
\multicolumn{3}{c|}{System (arithmetic)} & \multicolumn{2}{c}{Human}\\
\hline
\multicolumn{3}{p{3.3in}|}{\hspace{0.5cm}You are a helpful AI robot, you can solve mathematical problem accurately.\textbackslash nThe Math Word Problem(MWP), as a manifestation of questions, can be understood as the process of solving it by computing an operational expression. The following will provide a math word problem that you need to abstract into an arithmetic expression that can be directly interpreted by Python, such as 6 * (5+3).\textbackslash nYou can use functions from the \"math\"standard library.} & \multicolumn{2}{p{1.7in}}{**Math Word Problem**: \textbackslash n\{MWP\}\textbackslash n\textbackslash nYou need to provide your answer in the format of "<answer>...<\textbackslash answer>" at the end. You don't need to answer the MWP.}\\
\hline
\multicolumn{3}{c|}{System (equation)} & \multicolumn{2}{c}{Human}\\
\hline
\multicolumn{3}{p{3.3in}|}{\hspace{0.5cm}You are a helpful AI robot, you can solve mathematical problem accurately.\textbackslash nYou are a helpful AI robot, you can solve mathematical problem accurately, The Math Word Problem(MWP), as a manifestation of questions, can be understood as the process of solving a equation or system of equations. The following will provide a math word problem that you need to abstract into a equation or system of equations. Specifically, you need to first list the unknown variable(s) that need to be used after abstraction. If there are multiple unknown variables, use commas to separate them. Then list the abstract equation, and if there are multiple equations, list them in multiple lines.\textbackslash nYou can use functions from the \"math\"standard library.} & \multicolumn{2}{p{1.7in}}{**Math Word Problem**: \textbackslash n\{MWP\}\textbackslash n\textbackslash nYou need to provide your answer in the format of "<Unknown Variables>...<\textbackslash Unknown Variables>\textbackslash n<Equations> ...<\textbackslash Equations>"at the end.\textbackslash nYou don't need to answer the MWP.}\\
\hline
\multicolumn{3}{c|}{System (sorting)} & \multicolumn{2}{c}{Human}\\
\hline
\multicolumn{3}{p{3.3in}|}{\hspace{0.5cm}You are a helpful AI robot, you can solve mathematical problem accurately.\textbackslash nThe following will provide a math word problem(MWP) that you have to compare or sort some numbers in the MWP to solve it.\textbackslash nExtract the numbers that need to be compared or sorted from the questions.} & \multicolumn{2}{p{1.7in}}{**Math Word Problem**: \textbackslash n\{MWP\}\textbackslash n\{MWP\}\textbackslash n\textbackslash nYou need to provide your answer in the format of "<answer>...<\textbackslash answer>" at the end. You don't need to answer the MWP.}\\
\hline
\end{tabular}
\caption{Detailed description of subtask Question Abstraction.}
\end{table*}

\begin{table*}[htbp]
\centering
\begin{tabular}{ccccc}
\hline
\multicolumn{2}{c|}{\textbf{Subtask}} & \multicolumn{1}{c|} {\textbf{Aspect}} & \multicolumn{1}{c|}{\textbf{Multi-modal}} & \textbf{Size} \\ 
\hline
\multicolumn{2}{c|}{Task Decomposition (TD)} & \multicolumn{1}{c|}{Understanding} & \multicolumn{1}{c|}{No} & 4812\\ 
\hline\hline
\multicolumn{5}{c}{\textbf{Description}}\\ 
\hline
\multicolumn{5}{p{5in}}{\hspace{0.5cm}The LLMs are required to analyze the MWP and determine the necessary steps for solving it. The LLMs must have a sufficient understanding of the text and mathematical logic to answer correctly. This subtask has a certain degree of openness.}\\ 
\hline\hline
\multicolumn{5}{c}{\textbf{Prompt}}\\ 
\hline
\multicolumn{2}{c|}{System} & \multicolumn{3}{c}{Human}\\
\hline
\multicolumn{2}{p{1.7in}|}{\hspace{0.5cm}You are a helpful AI robot, you can solve mathematical problem accurately.\textbackslash nThe Math Word Problem(MWP), as a type of comprehensive mathematical problem, it may require various mathematical operations such as calculations and solving equations during to solve it.\textbackslash nBased on the MWP provided below, choose what mathematical operations are needed to solve it.} & \multicolumn{3}{p{3.3in}}{**Math Word Problem**: \textbackslash n\{MWP\}\textbackslash n\textbackslash n**Mathematical operation list**: \textbackslash nA: Additional information such as mathematical formulas, constants, theorems, etc. that are not directly provided in the question.\textbackslash nB: Solve an equation or system of equations.\textbackslash nC: Perform mathematical arithmetic.\textbackslash nD: Sort or compare the data in the question.\textbackslash nE: Identify and only identify the numbers in various formats provided in the information that are needed to solve the problem.\textbackslash nF: Identify the numerical unit(s) required to obtain the answer.\textbackslash nG: Identify and only identify the numbers in various formats provided in the image that are needed to solve the problem.\textbackslash nH: Quantify data in images that are not directly presented in numerical terms.\textbackslash nI: Count the number of certain objects in the picture. \textbackslash n\textbackslash nYou need to provide your answer in the format of "<answer>...<\textbackslash answer>" at the end. You only need to reply with the index letter. If multiple operations are selected, separate them with commas.\textbackslash nYou don't need to answer the MWP.}\\
\hline
\end{tabular}
\caption{Detailed description of subtask Task Decomposition.}
\end{table*}

\begin{table*}[htbp]
\centering
\begin{tabular}{ccccc}
\hline
\multicolumn{2}{c|}{\textbf{Subtask}} & \multicolumn{1}{c|} {\textbf{Aspect}} & \multicolumn{1}{c|}{\textbf{Multi-modal}} & \textbf{Size} \\ 
\hline
\multicolumn{2}{c|}{Math Knowledge (MK)} & \multicolumn{1}{c|}{Understanding} & \multicolumn{1}{c|}{No} & 58\\ 
\hline\hline
\multicolumn{5}{c}{\textbf{Description}}\\ 
\hline
\multicolumn{5}{p{5in}}{\hspace{0.5cm}This subtask evaluates the model's ability to leverage fundamental mathematical knowledge, such as approximations of constants and geometric formulas that are not explicitly provided in the problem. For example, the approximation of e or applying the quadratic formula for the root of an equation.}\\ 
\hline\hline
\multicolumn{5}{c}{\textbf{Prompt}}\\ 
\hline
\multicolumn{3}{c|}{System} & \multicolumn{2}{c}{Human}\\
\hline
\multicolumn{3}{p{3.3in}|}{\hspace{0.5cm}You are a helpful AI robot, you can solve mathematical problem accurately.\textbackslash nThe following will provide a math word problem(MWP). To solve this MWP, an additional knowledge point, such as a theorem or formula, is required. Please answer the name of this knowledge point.} & \multicolumn{2}{p{1.7in}}{**Math Word Problem**: \textbackslash n\{question\}\textbackslash n\textbackslash nYou need to provide your answer in the format of "<answer>...<\textbackslash answer>" at the end. You don't need to answer the MWP.}\\
\hline
\end{tabular}
\caption{Detailed description of subtask Math Knowledge.}
\end{table*}

\begin{table*}[htbp]
\centering
\begin{tabular}{ccccc}
\hline
\multicolumn{2}{c|}{\textbf{Subtask}} & \multicolumn{1}{c|} {\textbf{Aspect}} & \multicolumn{1}{c|}{\textbf{Multi-modal}} & \textbf{Size} \\ 
\hline
\multicolumn{2}{c|}{Arithmetic} & \multicolumn{1}{c|}{Calculation} & \multicolumn{1}{c|}{No} & 3532\\ 
\hline\hline
\multicolumn{5}{c}{\textbf{Description}}\\ 
\hline
\multicolumn{5}{p{5in}}{\hspace{0.5cm}This subtask evaluates the model's proficiency in performing basic mathematical expressions including operations such as four operations, root operation, exponential operation, etc.}\\ 
\hline\hline
\multicolumn{5}{c}{\textbf{Prompt}}\\ 
\hline
\multicolumn{3}{c|}{System} & \multicolumn{2}{c}{Human}\\
\hline
\multicolumn{3}{p{3.3in}|}{\hspace{0.5cm}You are a helpful AI robot, you can solve mathematical problem accurately.\textbackslash nPlease calculate the provided arithmetic expression.} & \multicolumn{2}{p{1.7in}}{**Arithmetic Expression**: \{expression\}\textbackslash n\textbackslash nYou need to provide your answer in the format of "<answer>...<\textbackslash answer>" at the end. If the result is a floating-point number, round to six decimal places.}\\
\hline
\end{tabular}
\caption{Detailed description of subtask Arithmetic.}
\end{table*}

\begin{table*}[htbp]
\centering
\begin{tabular}{ccccc}
\hline
\multicolumn{2}{c|}{\textbf{Subtask}} & \multicolumn{1}{c|} {\textbf{Aspect}} & \multicolumn{1}{c|}{\textbf{Multi-modal}} & \textbf{Size} \\ 
\hline
\multicolumn{2}{c|}{Equation Solving (EQ)} & \multicolumn{1}{c|}{Calculation} & \multicolumn{1}{c|}{No} & 1571\\ 
\hline\hline
\multicolumn{5}{c}{\textbf{Description}}\\ 
\hline
\multicolumn{5}{p{5in}}{\hspace{0.5cm}Require LLMs to solve both single-variable and systems of equations. It evaluates the model’s algebraic skills and its capacity for handling more advanced mathematical structures.}\\ 
\hline\hline
\multicolumn{5}{c}{\textbf{Prompt}}\\ 
\hline
\multicolumn{3}{c|}{System (equation)} & \multicolumn{2}{c}{Human}\\
\hline
\multicolumn{3}{p{3.3in}|}{\hspace{0.5cm}You are a helpful AI robot, you can solve mathematical problem accurately.\textbackslash n"Please solve the provided equation.} & \multicolumn{2}{p{1.7in}}{**Unknown Variable**: \{variable\}\textbackslash n**Equation**: \{equation\}"}\\
\hline
\multicolumn{3}{c|}{System (system of equations)} & \multicolumn{2}{c}{Human}\\
\hline
\multicolumn{3}{p{3.3in}|}{\hspace{0.5cm}You are a helpful AI robot, you can solve mathematical problem accurately.\textbackslash n"Please solve the provided system of equations.} & \multicolumn{2}{p{1.7in}}{**Unknown Variable**: \{variables\}\textbackslash n**System of Equations**: \{equations\}\textbackslash n\textbackslash nYou need to provide your answer in the format of "<answer>...<\textbackslash answer>" at the end. For example, "<answer>["x", "5"]<\textbackslash answer>". If the result is a floating-point number, round to six decimal places.}\\
\hline
\end{tabular}
\caption{Detailed description of subtask Equation Solving.}
\end{table*}

\begin{table*}[htbp]
\centering
\begin{tabular}{ccccc}
\hline
\multicolumn{2}{c|}{\textbf{Subtask}} & \multicolumn{1}{c|} {\textbf{Aspect}} & \multicolumn{1}{c|}{\textbf{Multi-modal}} & \textbf{Size} \\ 
\hline
\multicolumn{2}{c|}{Sorting} & \multicolumn{1}{c|}{Calculation} & \multicolumn{1}{c|}{No} & 582\\ 
\hline\hline
\multicolumn{5}{c}{\textbf{Description}}\\ 
\hline
\multicolumn{5}{p{5in}}{\hspace{0.5cm}This subtask evaluates a model's ability to arrange numbers or objects in a specific order, assesses its understanding of order relationships and computational reasoning.}\\ 
\hline\hline
\multicolumn{5}{c}{\textbf{Prompt}}\\ 
\hline
\multicolumn{3}{c|}{System} & \multicolumn{2}{c}{Human}\\
\hline
\multicolumn{3}{p{3.3in}|}{\hspace{0.5cm}You are a helpful AI robot, you can solve mathematical problem accurately.\textbackslash nPlease sort the following numbers in ascending order.} & \multicolumn{2}{p{1.7in}}{**Numbers**: \{numbers\}\textbackslash n\textbackslash nYou need to provide your answer in the format of "<answer>...<\textbackslash answer>" at the end.}\\
\hline
\end{tabular}
\caption{Detailed description of subtask Sorting.}
\end{table*}

\begin{table*}[htbp]
\centering
\begin{tabular}{ccccc}
\hline
\multicolumn{2}{c|}{\textbf{Subtask}} & \multicolumn{1}{c|} {\textbf{Aspect}} & \multicolumn{1}{c|}{\textbf{Multi-modal}} & \textbf{Size} \\ 
\hline
\multicolumn{2}{c|}{Formula Application (FA)} & \multicolumn{1}{c|}{Calculation} & \multicolumn{1}{c|}{No} & 246\\ 
\hline\hline
\multicolumn{5}{c}{\textbf{Description}}\\ 
\hline
\multicolumn{5}{p{5in}}{\hspace{0.5cm}This subtask requires the LLMs to recognize and apply specific formulas to solve problems and tests the LLMs' familiarity with mathematical relationships.}\\ 
\hline\hline
\multicolumn{5}{c}{\textbf{Prompt}}\\ 
\hline
\multicolumn{3}{c|}{System} & \multicolumn{2}{c}{Human}\\
\hline
\multicolumn{3}{p{3.3in}|}{\hspace{0.5cm}You are a helpful AI robot, you can solve mathematical problem accurately.\textbackslash nChoose the correct definition for the following theorem or formula.} & \multicolumn{2}{p{1.7in}}{**Theorem or Formula**: {formula}\textbackslash n**Options**:\textbackslash n{options}\textbackslash n\textbackslash nYou need to provide your answer in the format of "<answer>...<\textbackslash answer>" at the end.}\\
\hline
\end{tabular}
\caption{Detailed description of subtask Formula Application.}
\end{table*}

\begin{table*}[htbp]
\centering
\begin{tabular}{ccccc}
\hline
\multicolumn{2}{c|}{\textbf{Subtask}} & \multicolumn{1}{c|} {\textbf{Aspect}} & \multicolumn{1}{c|}{\textbf{Multi-modal}} & \textbf{Size} \\ 
\hline
\multicolumn{2}{c|}{Number Conversion (NC)} & \multicolumn{1}{c|}{Numerical Parsing} & \multicolumn{1}{c|}{No} & 3444\\ 
\hline\hline
\multicolumn{5}{c}{\textbf{Description}}\\ 
\hline
\multicolumn{5}{p{5in}}{\hspace{0.5cm}This subtask evaluates an LLM's ability to recognize and interpret important numbers in different formats, such as Arabic numerals, written words, and scientific notation. For example, "one hundred and three" or "1.13e+2" should be converted into "113". LLM also needs to avoid identifying invalid information.}\\ 
\hline\hline
\multicolumn{5}{c}{\textbf{Prompt}}\\ 
\hline
\multicolumn{3}{c|}{System} & \multicolumn{2}{c}{Human}\\
\hline
\multicolumn{3}{p{3.3in}|}{\hspace{0.5cm}You are a helpful AI robot, you can solve mathematical problem accurately.\textbackslash nThe Math Word Problem(MWP), as a type of comprehensive mathematical problem, it requires identify important information in the question to solve the problem.\textbackslash nThe following will provide a math word problem, and you need to identify and **ONLY** identify "the numbers in various formats provided in the information that are needed to solve the problem.} & \multicolumn{2}{p{1.7in}}{**MWP**: \{MWP\}\textbackslash n\textbackslash nYou need to provide your answer in the format of "<answer>...<\textbackslash answer>" at the end. You don't need to solve the MWP.}\\
\hline
\end{tabular}
\caption{Detailed description of subtask Number Conversion.}
\end{table*}

\begin{table*}[htbp]
\centering
\begin{tabular}{ccccc}
\hline
\multicolumn{2}{c|}{\textbf{Subtask}} & \multicolumn{1}{c|} {\textbf{Aspect}} & \multicolumn{1}{c|}{\textbf{Multi-modal}} & \textbf{Size} \\ 
\hline
\multicolumn{2}{c|}{Unit Conversion (UC)} & \multicolumn{1}{c|}{Numerical Parsing} & \multicolumn{1}{c|}{No} & 1093\\ 
\hline\hline
\multicolumn{5}{c}{\textbf{Description}}\\ 
\hline
\multicolumn{5}{p{5in}}{\hspace{0.5cm}In MWP, especially in physics-related problems, unit conversion is extremely important. This subtask measures an LLM's understanding of various units of measurement and its ability to convert between them. For example, converting "5 kW·h" to J or "100°C" to Fahrenheit.}\\ 
\hline\hline
\multicolumn{5}{c}{\textbf{Prompt}}\\ 
\hline
\multicolumn{3}{c|}{System} & \multicolumn{2}{c}{Human}\\
\hline
\multicolumn{3}{p{3.3in}|}{\hspace{0.5cm}You are a helpful AI robot, you can solve mathematical problem accurately.\textbackslash nThe Math Word Problem(MWP), as a type of comprehensive mathematical problem, it requires identify important information in the question to solve the problem.\textbackslash nThe following will provide a math word problem, and you need to identify the number with unit(s) required to solve the MWP. (Ignore numbers without units.)} & \multicolumn{2}{p{1.7in}}{**MWP**: \{MWP\}\textbackslash n\textbackslash nYou need to provide your answer in the format of "<answer>...<\textbackslash answer>" at the end. You don't need to solve the MWP.}\\
\hline
\end{tabular}
\caption{Detailed description of subtask Unit Conversion.}
\end{table*}

\begin{table*}[htbp]
\centering
\begin{tabular}{ccccc}
\hline
\multicolumn{2}{c|}{\textbf{Subtask}} & \multicolumn{1}{c|} {\textbf{Aspect}} & \multicolumn{1}{c|}{\textbf{Multi-modal}} & \textbf{Size} \\ 
\hline
\multicolumn{2}{c|}{Numeral Recognition (NR)} & \multicolumn{1}{c|}{Numerical Parsing} & \multicolumn{1}{c|}{Yes} & 714\\ 
\hline\hline
\multicolumn{5}{c}{\textbf{Description}}\\ 
\hline
\multicolumn{5}{p{5in}}{\hspace{0.5cm}This task assesses an LLM’s ability to extract mathematical content like numbers, variables, and formulas from images. LLM also needs to avoid identifying invalid information.}\\ 
\hline\hline
\multicolumn{5}{c}{\textbf{Prompt}}\\ 
\hline
\multicolumn{3}{c|}{System} & \multicolumn{2}{c}{Human}\\
\hline
\multicolumn{3}{p{3.3in}|}{\hspace{0.5cm}You are a helpful AI robot, you can solve mathematical problem accurately.\textbackslash nThe Visual Reasoning Problem(VRP), as a type of comprehensive mathematical problem, it requires identify important information in the image to solve the problem.\textbackslash nThe following will provide a visual reasoning problem and a image, you need to identify and **ONLY** identify the numbers in the image that are needed to solve the problem.} & \multicolumn{2}{p{1.7in}}{**VRP**: \{VRP\}\textbackslash n\textbackslash nYou need to provide your answer in the format of "<answer>...<\textbackslash answer>" at the end. You don't need to solve the VRP.}\\
\hline
\end{tabular}
\caption{Detailed description of subtask Numeral Recognition.}
\end{table*}

\begin{table*}[htbp]
\centering
\begin{tabular}{ccccc}
\hline
\multicolumn{2}{c|}{\textbf{Subtask}} & \multicolumn{1}{c|} {\textbf{Aspect}} & \multicolumn{1}{c|}{\textbf{Multi-modal}} & \textbf{Size} \\ 
\hline
\multicolumn{2}{c|}{Visual Data Quantification (VDQ)} & \multicolumn{1}{c|}{Numerical Parsing} & \multicolumn{1}{c|}{Yes} & 59\\ 
\hline\hline
\multicolumn{5}{c}{\textbf{Description}}\\ 
\hline
\multicolumn{5}{p{5in}}{\hspace{0.5cm}In the image, some data is not directly presented in numerical form, such as the time pointed by the clock or the length of an object. This subtask evaluates the model's ability to understand instructions and quantify nonvalue data in images.}\\ 
\hline\hline
\multicolumn{5}{c}{\textbf{Prompt}}\\ 
\hline
\multicolumn{3}{c|}{System} & \multicolumn{2}{c}{Human}\\
\hline
\multicolumn{3}{p{3.3in}|}{\hspace{0.5cm}You are a helpful AI robot, you can solve mathematical problem accurately.\textbackslash nIdentify the specified data from the following image. If the data is not presented directly in numerical form, you need to quantify it.} & \multicolumn{2}{p{1.7in}}{**Question**: \textbackslash n\{question\}\textbackslash n\textbackslash nYou need to provide your answer in the format of "<answer>...<\textbackslash answer>" at the end. You don't need to solve the VRP.}\\
\hline
\end{tabular}
\caption{Detailed description of subtask Visual Data Quantification.}
\label{tab: VDQ}
\end{table*}

\end{document}